\documentclass[sigconf, nonacm]{acmart}

\usepackage{makecell}
\usepackage{gensymb}

\begin{document}
\title{Machine Learning for Robotic Manipulation}

\author{Quan Vuong}
\affiliation{%
  \institution{UCSD}
}
\email{qvuong@ucsd.edu}

\begin{abstract}

The past decade has witnessed the tremendous successes of machine learning techniques in the supervised learning paradigm, where there is a clear demarcation between training and testing. In the supervised learning paradigm, learning is inherently passive, seeking to distill human-provided supervision in large-scale datasets into high capacity models. Following these successes, machine learning researchers have looked beyond this paradigm and became interested in tasks that are more dynamic. To them, robotics serve as an excellent test-bed, for the challenges of robotics break many of the assumptions that made supervised learning successful. Out of the many different areas within robotics, robotic manipulation has become a favorite area for researchers to demonstrate new algorithms because of the vast numbers of possible applications and its highly dynamical and complex nature. This document surveys recent robotics conferences and identifies the major trends with which machine learning techniques have been applied to the challenges of robotic manipulation.

\end{abstract}

\maketitle

\section{Introduction}

Robotic manipulation has long fascinated the research community \cite{toward_robotic_manipulation}, not only because it is an integrated science that requires inter-disciplinary knowledge, but also because a slew of primitive manipulation skills, such as throwing, using tools, and cooking, is so effortless for human to perform, yet so challenging to reproduce in machines. 

With the successes of deep learning in computer vision, machine learning practitioners have looked to robotics as the next frontiers of challenges. At the same time, roboticists have adopted machine learning machinery into their traditional workflows. Due to the need to perform sustained mechanical work on the world, robotics problems break the independent and identical distribution assumption of the supervised learning paradigm. These problems have motivated new approaches, both from machine learning and classical model-based perspectives. Looking at published material from major robotics conferences in the last year (CORL 2019-2020, ICRA 2020, RSS 2020, IROS 2020), this paper summarizes the recent trends in which machine learning has been applied to robotic manipulation.

We begin by reviewing the basic mathematical formalism and tools in \autoref{sec:background}. The subsequent sections identify distinct threads in which machine learning has been brought to bear on the challenges of robotic manipulations, as summarized below:

\begin{itemize}
    \item \autoref{sec:structure_for_learning}: Different forms of structure, such as graphs or 3D prior, can made learning more efficient and allow the integration of machine learning and traditional robotics methods.
    \item \autoref{sec:learning_to_improve_model_based_methods}: Learning can improve model-based methods when the ground truth physical models are hard to obtain or change during robot execution.
    \item \autoref{sec:data_collection_methods} and \autoref{sec:new_datasets}: New methods to collect datasets for robotics tasks and new benchmarks.
    \item \autoref{sec:learning_for_tactile}: Learning methods allow the use of high-dimensional, otherwise hard to write algorithms for, tactile signal.
\end{itemize}

Due to the inter-disciplinary nature of robotics research, there is not a commonly used standard set of notations. We thus refrain from describing the algorithms presented in recent papers with notations. When we discuss the background knowledge in \autoref{sec:background}, we use the notation convention often seen in the respective literature.

Please note that while we only discuss each paper in one section, the successful applications of machine learning methods to robotics problems often require successful execution in many orthogonal dimensions, i.e. both data collection and algorithm design. The inclusion of one paper in a particular section does not mean it does not have new insights to offer in other aspects.

\section{Background}
\label{sec:background}

In this section, we review the mathematical formalism and technical tools often found in robotics papers that use machine learning. We review both the machine learning tools and the more classical model-based methods. We divide the background material into the following sub-sections:

\begin{itemize}
    \item \autoref{sec:background_rl}: Reinforcement Learning.
    \item \autoref{sec:background_robot_kinematics}: Kinematics.
    \item \autoref{sec:background_robot_eq_of_motion}: Dynamics.
    \item \autoref{sec:background_control}: Control.
\end{itemize}

\subsection{Reinforcement Learning}
\label{sec:background_rl}

\begin{figure}
  \centering
  \includegraphics[width=0.8\linewidth]{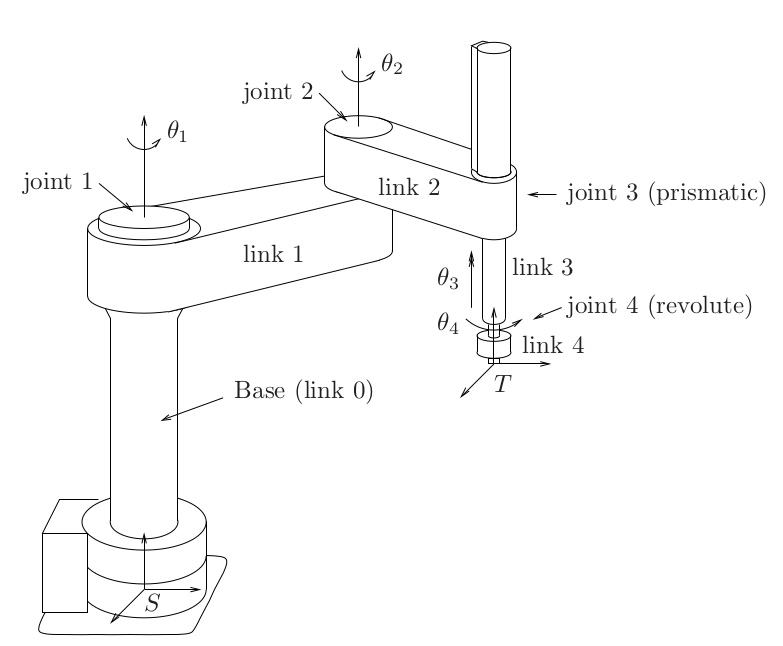}
  \caption{\textbf{An example of a simple manipulator from \cite{a_mathematical_introduction_to_robotic_manipulation}.} There are 4 joints, 3 of which are revolute (joint 1, 2, 4) and one prismatic joint (joint 3).}
  \label{fig:background_simple_manipulator}
\end{figure}

Reinforcement Learning is an abstraction of the real world that allows for optimization of long-term behavior. As a formalism, its main benefit comes from its generality, in which the robot designers specify the tasks the robots need to perform as the maximization of an expected sum of rewards, without making additional assumptions or having access to domain knowledge, such as the physical environments that the robots might be operating in. On the flip side, such tabula rasa approach means that the robots might require many interactions with their environment to accomplish the tasks. The application of Reinforcement Learning to robotics problems therefore often require substantial background engineering to maximize the amount of interactions the robots have with the environment. 

Typically, we represent an environment that the robot operates in and the task we want the robot to perform as a Markov Decision Process  $M = (\mathcal{S}, \mathcal{A}, T, T_0, R, H)$, with state space $\mathcal{S}$, action space $\mathcal{A}$, transition function $T$, initial state distribution $T_0$, reward function $R$, and horizon $H$. At each discrete timestep $t$, the robot is in a state $s_t$, picks an action $a_t$, arrives at $s'_t\sim T(\cdot|s_t, a_t)$, and receives a reward $R(s_t, a_t, s'_t)$. A policy is a function which maps a state to a distribution over the action space, from which we sample an action for the robot to take, $a_t \sim \pi(.|s_t)$. The definitions of the state space, action space, transition function, initial state distribution and reward function are problem-specific. However, the transition function is Markovian, wherein its output only depends on the current state $s_t$ and action $a_t$ and does not depend on the previous states and actions before the current time $t$. The main difference between Reinforcement Learning and classical robotic methods are the assumption of the lack of knowledge about both the transition function $T$ and the reward function $R$, where no knowledge of these functions are assumed and exploited in the design of algorithms.

Given a policy $\pi$ and a Markov Decision Process $M$, we can first sample an initial state from the initial state distribution $s_0 \sim T_0$, use the policy to pick the first action $a_0$, which incurs a reward $r_0 = R(s_0, a_0, s_1)$ and transitions the Markov Decision Process to the next state $s_1 = T(s_0, a_0)$. If we repeat the procedure for $H$ times, we obtain a trajectory: 
$$\tau_M = (s_0, a_0, r_0, s_1, a_1, r_1, \ldots, s_{H-1}, a_{H-1}, r_{H-1}, s_H)$$ 

The performance measure of policy $\pi$ is the expected sum of rewards:
$$J_M(\pi) = \mathbb{E}_{\tau_M \sim \pi}[\sum_{t=0}^{H-1} R(s_t, a_t, s'_t)]$$ 
where the expectation is over the stochasticity present in the transition function, the reward function and the policy.

In Deep Reinforcement Learning, we parameterize the policy $\pi$ with a deep neural network. Many Deep Reinforcement Learning algorithms have been proposed and are now routinely used in robotic papers published at major robotics conferences. In addition to the policy, another major object often found in Reinforcement Learning algorithms is the state-value function $Q_\pi(s_{t^{''}}, a_{t^{''}})$ which approximates the value of taking action $a_{t^{''}}$ at state $s_{t^{''}}$ and thereby picking action by sampling from policy $\pi$ until the end of the trajectory. That is: $$Q_\pi(s_{t^{''}}, a_{t^{''}}) \approx  \mathbb{E}_{\pi, M}[\sum_{t=t^{''}}^{H-1} R(s_t, a_t, s'_t) ].$$

Together with collaborators, I have contributed five Deep Reinforcement Learning algorithms as publications at ICLR 2019 \cite{supervised_policy_update} (proposing a new algorithm based on convex optimization theory), neuRIPS 2019 spotlight \cite{better_exploration_with_optimistic_actor_critic} (improving sample efficiency with optimistic exploration), ICML 2020 \cite{towards_simplicity_in_deep_rl_streamlined_off_policy_learning} (understanding state-of-the-art method), neuRIPS 2020 \cite{multi_task_batch_rl_with_metric_learning} (extending Reinforcement Learning formalism to the multi-task offline setting) and neuRIPS 2020 spotlight \cite{first_order_constrained_optimization_in_policy_space} (Safe Reinforcement Learning algorithm).

Reinforcement Learning algorithms are usually presented absent of any assumptions about the environment in which the algorithms are deployed. In practice, when Reinforcement Learning algorithms are applied to robotics problem, some amount of domain expertise are assumed, such as an understanding of the mechanical properties of the robots or a low-level controller to convert desired torque to electrical signal at the motor-level.

\subsection{Robot Kinematics}
\label{sec:background_robot_kinematics}

Commonly used manipulators in research setting consist of a set of rigid links connected together as an acyclic graph by a set of joints. We attach motors to the joints so that we can generate motion of the links to perform tasks. The motion that each joint affords typically corresponds to subset of the special Euclidean group $SE(3)$, such as translation for prismatic joint or rotation for revolute joint. The kinematics of a manipulator describes the relationship between the motion of the joints of the manipulator and the resulting motion of the rigid links. 

\autoref{fig:background_simple_manipulator} illustrates the schematic of a simple manipulator with 4 joints and 5 links. Associated with each joint is its current joint angle $\theta_i$. The vector of all joint angles $\theta = [\theta_1, \theta_2, \theta_3, \theta_4]$ represent the generalized coordinates of the manipulator. These are called ``generalized" because even though they do not specify the positions of all points on the manipulator, knowing their values allow us to compute the position of any point because of the rigid body assumptions of the links. There is a stationary coordinate frame $S$, commonly referred to as the base frame. At the end of the manipulator is another frame $T$, commonly referred to as the tool frame. As the joint value changes, the configuration of frame $T$ with respect to frame $S$ changes. The forward kinematics problem determines the configuration of frame $T$ with respect to frame $S$ given the values of the joint angles. The forward kinematics problem has an elegant closed form solution called the product of exponentials formula.

Let $g_{st}(\theta)$ represents the configuration of frame $T$ with respect to frame $S$ for a particular value of the joint angles $\theta$. For each joint as indexed by $i$, construct a unit twist $\hat{\xi}_i$, corresponding to the axis of the screw motion of the $i$-th joint with all remaining joint angles held fixed at $0$. The product of exponentials formula for the forward kinematics problem for a manipulator with $n$ joint is as followed:
$$
g_{s t}(\theta)=e^{\widehat{\xi}_{1} \theta_{1}} e^{\widehat{\xi}_{2} \theta_{2}} \cdots e^{\widehat{\xi}_{n} \theta_{n}} g_{s t}(0)
$$
The inverse problem to the forward kinematics problem is the inverse kinematics problem: given a desired configuration of the frame $T$ with respect to the frame $S$, we wish to find the joint angles which achieve that configuration. The inverse kinematics problem is substantially harder than the forward kinematics problem and can have zero, one or many solutions depending on the number of degrees of freedom of the manipulator. There are two general solution families to the inverse kinematics problem. For simple problems, we can decompose the problem into easier sub-problems by hand, such as solving for a rotation such that a point is rotated to another point given a fixed rotation axis. There are also mature software library to solve the inverse kinematics problem, such as IKFast from OpenRave. 

Taking the time derivative of the configuration of $T$ with respect to $S$, we obtained $J_{s t}^{s} \in R^{6 \times n}$, referred to as the spatial manipulator Jacobian. The spatial manipulator Jacobian is of central importance in classical model-based robotics methods. It relates the velocity of the joint angles $\dot{\theta}$ and the velocity of the frame $T$ with respect to frame $S$, as seen from the $S$ frame, $V_{s t}^{s}$:
$$
V_{s t}^{s}= J_{s t}^{s}(\theta) \dot{\theta}
$$
The transpose of the spatial manipulator Jacobian also allows us to relate a spatial wrench $F_s$ applied at the frame $T$ and the corresponding joint torques:
$$
\tau=\left(J_{s t}^{s}\right)^{T} F_{s}
$$
There are also body-frame equivalent of the base-frame spatial manipulator Jacobian, which relates the joint velocity to end-effector velocity and joint torque to end-effector wrench as seen from the body frame.

\subsection{Robot Equation of Motion}
\label{sec:background_robot_eq_of_motion}

While kinematics allows us to relate different quantities at a particular point in time, robotics is sequential in nature. The study of dynamics and more specifically the robot equation of motion allow us to relate different properties of the robots as they evolve in time.

The robot equation of motion, which relates the joint positions (angles) $\theta$, velocities  $\dot{\theta}$, accelerations $\ddot{\theta}$ and the torques at the joints can be derived from Lagrange's equations:
$$
\frac{d}{d t} \frac{\partial L}{\partial \dot{q}_{i}}-\frac{\partial L}{\partial q_{i}}=\Upsilon_{i} \quad i=1, \ldots, m
$$
where $q_i$ can be thought of as the angle of joint $i$, $L(q, \dot{q})=T(q, \dot{q})-V(q)$ is the Lagrangian (kinetic minus potential energy of the system), $m$ is the number of joints and $\Upsilon_{i}$ is the external force acting on joint $i$. For an open-chain manipulator, we can expand Lagrange's equation to take into account the mass distribution of the rigid links and arrive at the following manipulator equation:
$$
M(\theta) \ddot{\theta} + C(\theta, \dot{\theta}) \dot{\theta}+N(\theta, \dot{\theta})=\tau
$$
where $M$ is the mass matrix, $C$ is the Coriolis matrix, $N$ models gravity terms and other external forces which act at the joints and $\tau$ indicates the torque exerted by the actuator at the joints. $M$ is always symmetric and positive definite. $\dot{M} - 2C$ is a skew-symmetric matrix. $M, C, N$ are often obtained by careful experimental modeling of the robots.

\begin{figure}
  \centering
  \includegraphics[width=\linewidth]{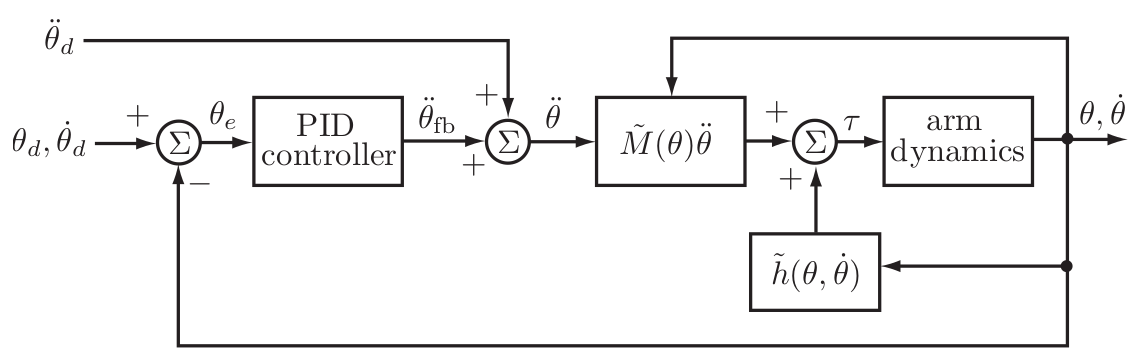}
  \caption{\textbf{A graphical illustration of computed torque controller from \cite{modern_robotics_mechanics_planning_and_control}.} The desired sequence of joint angles $\theta$ and its derivatives are combined with a model of the dynamics $\tilde{M}, \tilde{h}$ to compute the torque input into the robot low-level controller.}
  \label{fig:background_computed_torque_controller}
\end{figure}

\subsection{Control}
\label{sec:background_control}

Given an understanding of how the robot evolves in time, we are now interested in commanding torques at the joints such that the robot performs useful mechanical work.

To control the manipulator, we can specify a desired sequence of joint angles $\theta_d(t)$ indexed by time that the robot's joints should closely follow. A popular choice of control input is torque input because it allows for the use of a dynamical model of the robot in the design of the control law \cite{modern_robotics_mechanics_planning_and_control}. Let: 
$$
h(\theta, \dot{\theta}) = C(\theta, \dot{\theta}) \dot{\theta}+N(\theta, \dot{\theta}).
$$
Let the controller's model of the dynamics be:
$$
\tau=\tilde{M}(\theta) \ddot{\theta}+\tilde{h}(\theta, \dot{\theta})
$$
where the model is perfect if $\tilde{M}(\theta)=M(\theta)$ and $\tilde{h}(\theta, \dot{\theta})=h(\theta, \dot{\theta})$. Given a twice-differentiable desired sequence of joint angles $\theta_d(t)$, we can write the feed-forward torque controller as followed:
$$
\tau(t) = \tilde{M}\left(\theta_{d}(t)\right) \ddot{\theta}_{d}(t) + \tilde{h}\left(\theta_{d}(t), \dot{\theta}_{d}(t)\right)
$$
If the controller's model of the dynamics is perfect and there is no initial error in the joint angles, then we expect the feed-forward torque controller to command the joint torques such that the joint angles follow the desired sequence exactly.

However, the feed-forward controller will not perform well if the model of the dynamics is inaccurate or if there are initial errors in the joint angles. Define the joint error as:
$$
\theta_e(t) = \theta_d(t) - \theta(t)
$$
We can use the error as a signal to change the commanded torque such that the robot tracks the desired sequence of joint angles better over time, leading to the following popular computed torque controller:
$$
\tau=\tilde{M}(\theta)\left(\ddot{\theta}_{d}+K_{p} \theta_{e}+K_{i} \int \theta_{e}(\mathrm{t}) d \mathrm{t}+K_{d} \dot{\theta}_{e}\right) + \tilde{h}(\theta, \dot{\theta})
$$
where $K_p$, $K_i$, and $K_d$ are positive-definite matrices and can be chosen to achieve good transient error response. This controller is also known as the feed-forward plus feedback linearizing controller or the inverse dynamics controller. This controller uses the planned acceleration $\ddot{\theta}_d$ to generate non-zero torque command even when the error in joint angles is zero. It is called feedback linearizing because the error signal from $\theta$ and $\dot{\theta}$ are used to generate a linear error dynamics that ensures exponential decay of the joint error. The term $\tilde{h}(\theta, \dot{\theta})$ cancels the dynamics that depends non-linearly on the state of the system $\theta, \dot{\theta}$. The mass matrix $\tilde{M}(\theta)$ converts the desired joint accelerations into commanded joint torques. The commanded joint torques are sent to low-level controller at the joint to convert them into corresponding electrical signal. \autoref{fig:background_computed_torque_controller} illustrates the computations involved in the computed torque controller.

To generate the desired joint sequence $\theta_d(t)$ to achieve a specific task, such as opening a door, we can use a high-level motion planner. Given a model of the robot's environment and the desired configuration of the environment, the planner computes a sequence of joint angles such that by following the sequence, the robot accomplishes the desired change in environment's configuration. For robotic manipulation tasks, it is often more convenient to specify the motion of the robot to achieve a particular task as a sequence of configurations of the end-effectors. There are two potential choices to convert the desired sequence of configurations of the end-effectors into joint torques: either use inverse kinematics to convert the desired sequence of end-effector configurations into desired sequence of joint angles, or implement the equivalent of the controller laws above natively in end-effector space, rather than joint space. 

\begin{figure}
  \centering
  \includegraphics[width=\linewidth]{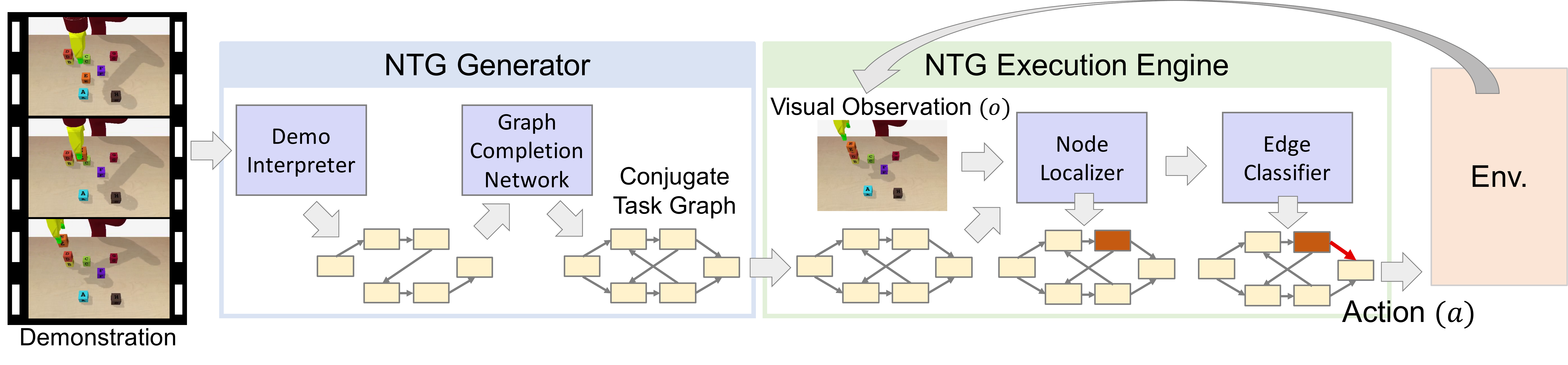}
  \caption{\textbf{Conjugate Task Graph system from \cite{neural_task_graphs_generalizing_to_unseen_tasks_from_a_single_video_demonstration}. } The Demo Interpreter generates the Conjugate Task Graph (CTG) from the demonstration video. The Graph Completion Network completes the CTG by filling in the missing edges. The model executes the task graph by associating the current visual observation with a node in the graph with the Node Localizer and choose action to take with the Edge Classifier.}
  \label{fig:structure_for_learning_ntg}
\end{figure}

The controllers above implement motion control, where we specify the desired motion of the robot manipulator in joint space. In addition to specifying the motion of the robot, we can also specify the forces with which the robot end-effectors should exert on its environment. There are also hybrid force-motion controllers that allow the robot to exert motion in some axis while exerting force in others. Another popular class of controller is impedance controllers, which can be thought of as ensuring that the robot responses as a virtual spring when it comes into contact with the environment, thus providing a degree of flexibility and safety. Typically, the choice of a good controller depends on the specific tasks we want the robots to accomplish.

\section{Adding explicit structure to aid learning}
\label{sec:structure_for_learning}

In this section, we discuss the plethora of ways in which researchers have used additional structure and insights to make learning more successful in robotic manipulation. We organize the material as followed:

\begin{itemize}
    \item \autoref{sec:structure_for_learning_graph}: Graph inductive bias.
    \item \autoref{sec:structure_for_learning_hier_structure}: Hierarchical structure.
    \item \autoref{sec:structure_for_learning_3D}: 3D prior.
    \item \autoref{sec:structure_for_learning_act_param}: Different forms of action parameterization.
    \item \autoref{sec:structure_for_learning_language}: Integration of language.
\end{itemize}

\subsection{Graph Inductive Bias}
\label{sec:structure_for_learning_graph}

Graphical data structures allow algorithm designers to specify relationships and hierarchy between abstract concepts in intuitive ways. It is thus of no surprise that they have found applications in object manipulation tasks, where the algorithms have to reason about the spatial and temporal relationships of objects. \cite{graph_structured_visual_imitation, towards_practical_multi_object_manipulation_using_relational_rl, neural_task_graphs_generalizing_to_unseen_tasks_from_a_single_video_demonstration} use the hierarchical and relational structure of graphs to improve the generalization capability of neural networks. 

Task graph is a common representation of high-level execution plans in robotics, where the states are nodes in the graph and the edges represent actions that transitions the environment from one state to another. \cite{neural_task_graphs_generalizing_to_unseen_tasks_from_a_single_video_demonstration} consider the problem of visual imitation, where the states are images. They thus assume their action space is a discrete, finite set and represent a task by its conjugate task graph (CTG), where the actions are the nodes and the states are the edges. Their assumption of a small and discrete action space allows them to keep the size of the CTG relatively small, which would not have been easily doable if the nodes in the task graph represent visual states instead. Their algorithm consists of two phases: CTG generation from a demonstration video and CTG execution (\autoref{fig:structure_for_learning_ntg}). To generate the CTG, the model first predicts the action sequence executed in the video, which forms the node in the CTG. The nodes are connected based on their order of execution in the predicted action sequence. The model then completes the CTG by predicting missing possible transitions between nodes in the CTG. Given the generated CTG, the model executes the corresponding task by localizing which nodes in the CTG corresponds to the current visual observation. Given the localized node, it then predicts which node (action) connected to the localized node in the CTG should be executed next. In terms of dataset, the method proposed in \cite{neural_task_graphs_generalizing_to_unseen_tasks_from_a_single_video_demonstration} requires videos of the same task being accomplished with different action sequences. The action sequences should also be temporally aligned with the frames in the videos. These two sources of supervision provides the necessary signal for the CTG completion step, which predicts possible transitions between actions not seen in a demonstration video.

\begin{figure}
  \centering
  \includegraphics[width=\linewidth]{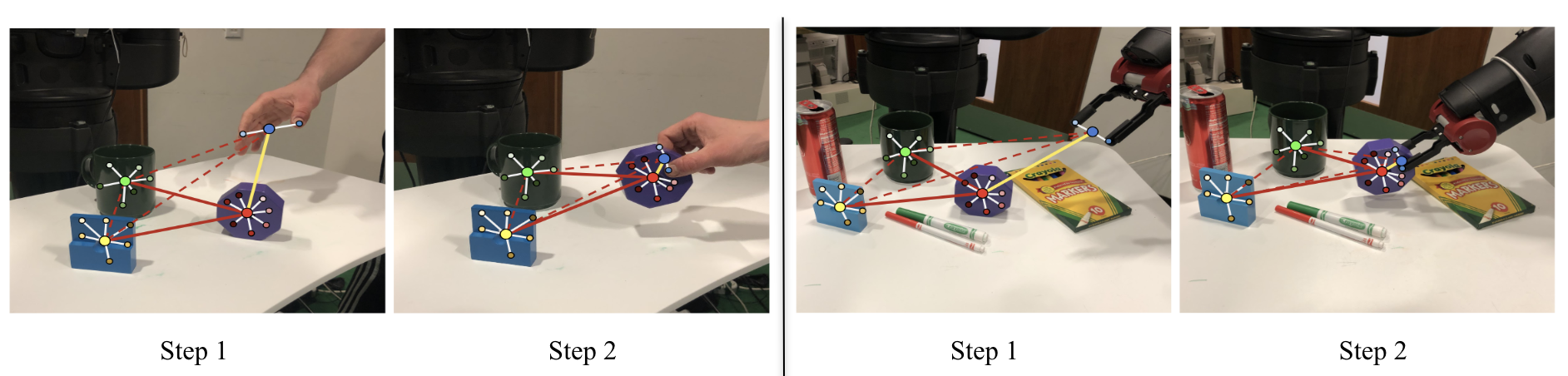}
  \caption{\textbf{An example of the hierarchical graph video representation from \cite{graph_structured_visual_imitation}.} Given the changes in the relative spatial arrangement of visual entities in the human demonstration from Step 1 to 2 (first two images), the robot detects matching visual entities in its workspace and moves to achieve similar changes in the entities' relative spatial arrangement (last two images).}
  \label{fig:structure_for_learning_graph_visual_imitation}
\end{figure}

\begin{table*}[t]
  \caption{An overview of how recent works design hierarchy or modularized components.}

  \label{tab:structured_for_learning_overview}
      
  \begin{tabular}{ccc}
    \toprule
    Papers & High-level component & Low-level component \\
    \midrule
    \cite{a_long_horizon_planning_framework_for_manipulating_rigid_pointcloud_objects} & Sampling-based planner & \makecell{Learned sampler to sample sub-goal \\ (desired object pose) \\ for the sampling-based planner to plan with} \\
    \midrule
    \cite{dynamics_learning_with_cascaded_variational_inference_for_multi_step_manipulation} & \makecell{Model Predictive Controller (MPC) planner \\ planning in the space of sub-goal} & \makecell{MPC planner planning low-level motion to achieve sub-goal} \\
    \midrule
    \cite{self_supervised_sim_to_real_adaptation_for_visual_robotic_manipulation} & Learned representation adapted from sim2real &  \makecell{Learned controller \\ input: the learned representation\\output: desired robot velocity} \\
    \midrule
    \cite{efficient_bimanual_manipulation_using_learned_task_schemas} & \makecell{Sequences of parameterized skills, \\such as go-to pose}  & Parameters of the skills \\
    \midrule
    \makecell{\cite{, learning_to_combine_primitive_skills_a_step_towards_versatile_robotic_manipulation, iris_implicit_reinforcement_without_interaction_at_scale_for_learning_control_from_offline_robot_manipulation_data, hrl4in_hierarchical_rl_for_interactive_navigation_with_mobile_manipulators} \\ 
    \cite{sim_to_real_transfer_of_bolting_tasks_with_tight_tolerance, learning_to_compose_hierarchical_object_centric_controllers_for_robotic_manipulation, learning_hierarchical_control_for_robust_in_hand_manipulation, controlling_contact_rich_manipulation_under_partial_observability}}
     & \makecell{Reinforcement Learning controller \\output: parameters of low-level controllers or \\ their composition or sequence} & \makecell{Manually-defined or learned controllers.\\For example, trajectory tracking controller or \\ Position, force or torque controllers or \\ Imitation Learning controllers} \\
    \midrule
    \cite{learning_reactive_motion_policies_in_multiple_task_spaces_from_human_demonstrations} & Riemannian motion policies flow controller & Imitation Learning controller \\
    \midrule
    \cite{task_conditioned_variational_autoencoders_for_learning_movement_primitives} & \makecell{Manually specified task-parameters. \\For example: where to pour } & \makecell{Learned parameters indicating how the tasks should be done.\\For example: how long to pour for} \\
    \bottomrule
  \end{tabular}
  
\end{table*}

Instead of representing the policy as factorized modules that are trained separately as in \cite{neural_task_graphs_generalizing_to_unseen_tasks_from_a_single_video_demonstration}, in visual Reinforcement Learning, the policy is commonly parameterized by convolutional networks followed by fully connected layers and trained end-to-end. 
While such parameterization suffices to achieve high single-task performance, \cite{towards_practical_multi_object_manipulation_using_relational_rl} claims that it impedes generalization in multi-object manipulation tasks. 
Instead, they represent their policies by attention-based graph neural network (GNN), which has been shown to generalize better than convolutional networks for structured data \cite{graph_neural_networks_a_review_of_methods_and_applications}. \cite{towards_practical_multi_object_manipulation_using_relational_rl} trains the GNN-parameterized policies to perform block-stacking tasks through an increasingly more challenging curriculum where the numbers of block increase over time. 
The trained policy can generalize without further training to block-stacking tasks where the number of blocks that needed to be stacked are not seen during training. 
The use of GNN also allows some degree of interpretation into what the policy has learnt.
Each node in the GNN represents a block in the environment. 
By examining the attention weights between the nodes, the author claims that the policy pays more attention to the sub-set of blocks relevant to solving a sub-problem, such as stacking one particular block onto another. They thus hypothesize that the policy has learnt to decompose the complex multiple-block stacking tasks into simpler sub-problem of stacking one block at a time. 

Compared to \cite{neural_task_graphs_generalizing_to_unseen_tasks_from_a_single_video_demonstration}, \cite{graph_structured_visual_imitation} proposes a different method to solve visual imitation. 
They propose hierarchical graph video representation.
In this graph, the nodes represent visual entities detected in the scenes, such as keypoints on objects, and the edges represent their relative 3D spatial arrangement. 
Given a demonstration video, they first construct the corresponding hierarchical graph representation. They then compare the relative arrangement of the visual entities in the scene between the demonstration video and the imitator's workspace to construct a reward function. They use this reward function for trajectory optimization to generate actions to be executed on the imitating robot (\autoref{fig:structure_for_learning_graph_visual_imitation}). To detect keypoints on objects, match keypoints across the demonstration-imitator view and across time in a self-supervised manner, they perform multi-view self-supervised point feature learning, which is explained in more details in \autoref{sec:structure_for_learning_3D}. 
\cite{graph_structured_visual_imitation} demonstrates better generalization compared to method which maps the input images directly into a high-dimensional embedding space and use distance in the embedding space as the reward function.
However, it is unclear how to extend their methods to more complex scenes where objects can not be represented as keypoints, such as in water pouring tasks.

A common limitation among methods which integrate graphs into learning is the data requirement. The dataset must provide relational signal to allow the learning process to capture the hierarchical and relational structure in the problem. For example, the method in  \cite{neural_task_graphs_generalizing_to_unseen_tasks_from_a_single_video_demonstration} performs worse than baselines when there are less than $100$ training tasks. For each task, the method also requires demonstrations that accomplish the tasks with different action sequences, which is crucial to learn the action-to-action transition and train the Conjugate Task Graph completion network. Similarly, the use of attention-based graph neural network to parameterize the policy in  \cite{towards_practical_multi_object_manipulation_using_relational_rl} only demonstrates benefit over simple multi-layer perceptron parameterization when the policies are trained with the curriculum, which might not be realizable in practice on real robots. These two methods are also not demonstrated on real robots.

\subsection{Hierarchical structure}
\label{sec:structure_for_learning_hier_structure}

Previous applications of machine learning to robotic manipulation \cite{deep_rl_for_robotic_manipulation, end_to_end_training_of_deep_visuomotor_policies, solving_rubiks_cube_with_a_robot_hand} take a black-box approach where the controller is a deep neural network, trained in an end-to-end manner, with no explicit hierarchical structure. 
A major trend in recent robotic manipulation research is the explicit integration of hierarchy into algorithm design, which allows for the integration of machine learning and classical model-based techniques. 
\autoref{tab:structured_for_learning_overview} provides an overview of how recent works design the hierarchical structures. In this section, I discuss three representative works, where the high and low level components are both learned \cite{iris_implicit_reinforcement_without_interaction_at_scale_for_learning_control_from_offline_robot_manipulation_data}, the high-level controller is learned but the low-level controllers are designed using model-based method \cite{learning_hierarchical_control_for_robust_in_hand_manipulation}, the low-level controllers are learned and combined into a high-level controller with model-based method \cite{learning_reactive_motion_policies_in_multiple_task_spaces_from_human_demonstrations}.

\begin{figure}
  \centering
  \includegraphics[width=\linewidth]{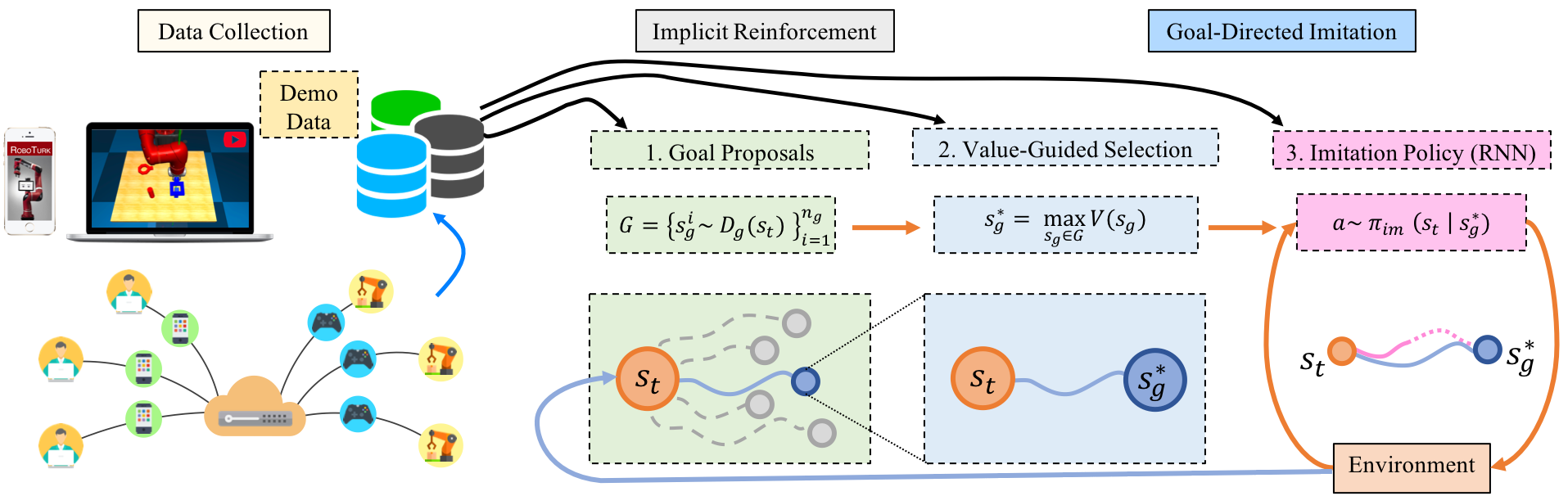}
  \caption{\textbf{Overview of system to learn from crowd-sourced demonstrations from \cite{learning_reactive_motion_policies_in_multiple_task_spaces_from_human_demonstrations}. } The paper uses the demonstration data to train the Goal Proposal and Value-Guided Selection modules, which together comprise the high-level goal selection mechanism. They also use the demonstration data to train a low-level Imitation Policy to imitate short sequences in the demonstration to achieve a goal.}
  \label{fig:structure_for_learning_iris}
\end{figure}

\cite{iris_implicit_reinforcement_without_interaction_at_scale_for_learning_control_from_offline_robot_manipulation_data} tackles the challenges of learning policies from existing large-scale and crowd-sourced demonstrations. There are two properties of the dataset which make learning challenging: sub-optimality and diversity. Since the demonstrations are collected from many different human workers, different workers might have different strategies of controlling the robot to perform the same tasks. Such multi-modal demonstration is difficult for naive Imitation Learning method since for the same state, there might be multiple different good actions to take. The method tackles this challenge through hierarchy: a high-level goal selection module and a low-level goal-conditioned imitation controller.  \autoref{fig:structure_for_learning_iris} provides a high-level overview of the different components in the method. 
The high-level goal selection module selects goal states for the low-level controller to move towards. 
The high-level goal selection module consists of 2 components: a conditional Variational Auto-Encoder (cVAE) which proposes goal states and a value function that models the expected return of the goal states. 
Given the current state, the cVAE proposes multiple future goal states. 
The value function evaluates them and sets the state with the highest expected return as the goal state for the low-level controller. 
The paper trains the cVAE using the evidence lower bound to capture the distributions of states within the demonstration, thereby modeling the multi-modal aspects of the demonstrations. 
They also train the value function using \cite{off_policy_deep_rl_without_exploration} to tackle the value function divergence issues often seen when Reinforcement Learning algorithms are trained from purely existing data without further interactions with the environment. 
They train the low-level controller using standard supervised Behavior Cloning loss to reach a goal state given the current state. 
They choose the goal state such that getting from the current state to the goal state does not require executing many actions. 
The paper argues that the multi-modal aspects of the demonstration only manifests over temporally extended sequences of state-action and is not present in shorter state-action sequences. The low-level controller thus does not need to tackle the multi-modality issue.

\begin{figure}
  \centering
  \includegraphics[width=0.7\linewidth]{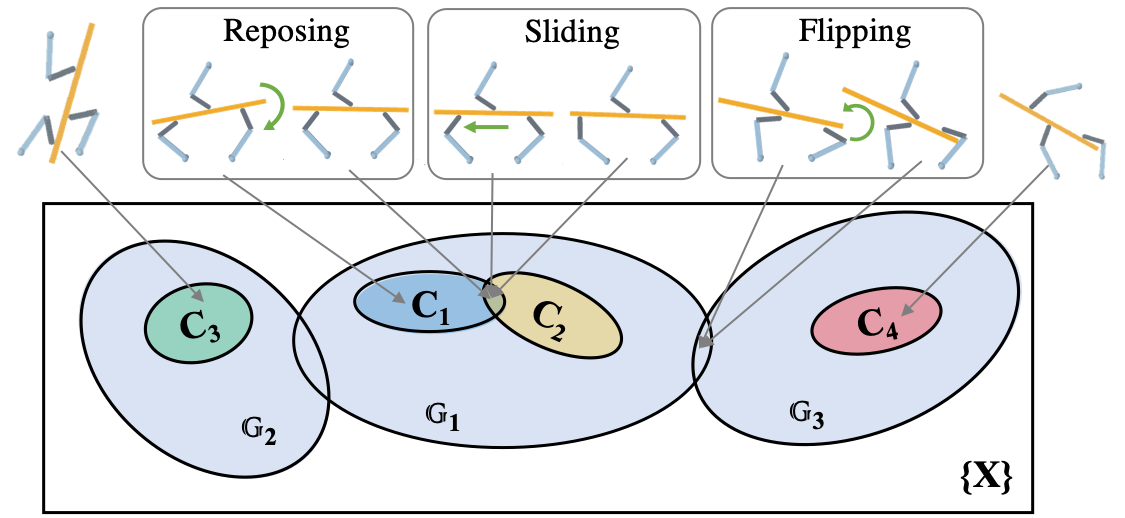}
  \caption{\textbf{Overview of the pose space from \cite{learning_hierarchical_control_for_robust_in_hand_manipulation}. } Each contact configuration $C$ allows the fingers to move the object to a limited range of poses within the entire pose space of the object $X$. Thus, to move the object from an initial pose to a desired pose, the fingers might have to change the contact configuration using either the Sliding or Flipping motion primitives.}
  \label{fig:structure_for_learning_hierarchical_controller_in_hand_manipulation}
\end{figure}

Instead of learning both the high and low level components, \cite{learning_hierarchical_control_for_robust_in_hand_manipulation} only learns the high-level controller while the low-level controllers are designed using model-based methods. \cite{learning_hierarchical_control_for_robust_in_hand_manipulation} tackles in-hand manipulation of objects. 
In-hand manipulation is the task of changing the pose of an object while ensuring the object and the manipulator maintain continuous contact. 
The manipulator considered consist of three fingers on a planar surface, assuming hard contact model and precision grasp. The object is an planar bar. 
Each contact configuration of the fingers with the object allow the fingers to move the object to a range of poses, but not all possible poses of the object. 
Thus, to move an object from an initial pose to a desired pose, the fingers might have to change contact configuration (\autoref{fig:structure_for_learning_hierarchical_controller_in_hand_manipulation}). 
There are three low-level controllers which are designed assuming access to the object models and using techniques discussed in \autoref{sec:background_robot_eq_of_motion} and \autoref{sec:background_control}: Reposing, Sliding and Flipping. 
Reposing allows the fingers to change the pose of the object while maintaining the current contact configuration. 
Sliding and Flipping change the contact configuration. 
To determine which low-level controller to use in a particular state, the paper learns a Reinforcement Learning controller which chooses which low-level controller to use and picks the values of its parameters. 

\begin{figure}
  \centering
  \includegraphics[width=0.5\linewidth]{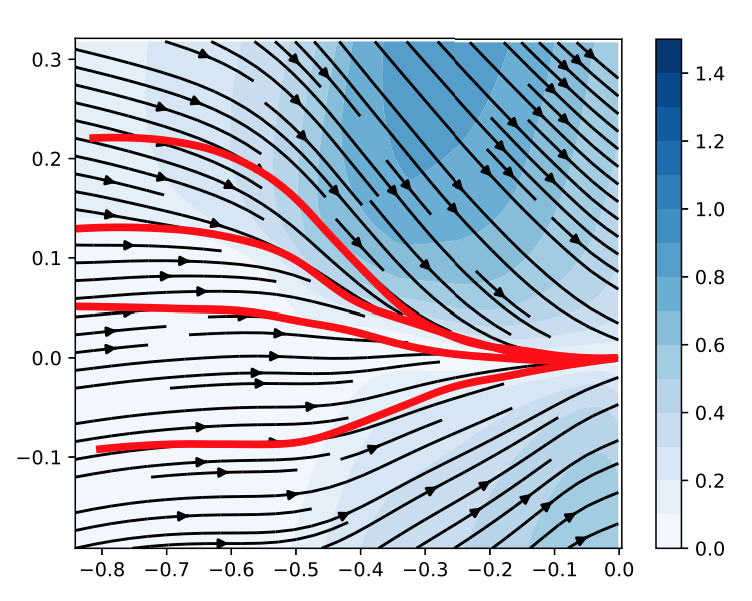}
  \caption{\textbf{Illustration of the learnt Riemannian Motion Policy from \cite{learning_reactive_motion_policies_in_multiple_task_spaces_from_human_demonstrations}. } The red curves represent the human demonstrations. The learnt vector field generates desired acceleration such that the human demonstrations can be reproduced.}
  \label{fig:structure_for_learning_RMP}
\end{figure}

Lastly, we can also learn the low-level controllers and combine them using a manually specified high-level mechanism. RMPflow \cite{learning_reactive_motion_policies_in_multiple_task_spaces_from_human_demonstrations} is a motion generation framework which decomposes a task into multiple sub-tasks, each with its own sub-task space. For example, a manipulator might need its end-effector to reach a particular point in 3D space while avoiding collision, a common problem for object manipulation in clutter. The basic object in the RMPflow framework is the Riemannian Motion Policy (RMP), which specifies the desired acceleration to accomplish one sub-task. 
Given multiple RMPs, a RMP-tree combines the accelerations generated by the sub-tasks into the desired acceleration in joint space, with the contribution from each sub-task weighted by its own geometric metric. 
One important feature of the RMPflow framework is that it preserves the stability property of the sub-task policies: if the sub-task policy is a virtual dynamical system of a particular form \cite{rmpflow_a_computational_graph_for_automatic_motion_policy_generation}, then the combined policy is also of this form and is stable in the sense of Lyapunov. Previous works on RMPflow manually specify the sub-task RMP policies, thereby limiting the framework to motion whose policies are easily specified or known a prior. 
They thus learn the sub-task RMP policies using Imitation Learning of human demonstration. 
The learnt RMP is parameterized by a potential function and a position-dependent metric matrix. 
They compute the potential function in a parameter-free manner as the weighted convex combination of simple exponential potential functions, with each directed toward a point along a nominal human demonstration trajectory. They train a neural network to take as input the current state and output the metric matrix. 
The metric matrix wraps the direction of the gradient of the potential function such that the final vector field generates desired acceleration that reproduces multiple human demonstrations (\autoref{fig:structure_for_learning_RMP}).

\subsection{3D prior}
\label{sec:structure_for_learning_3D}

3D geometric prior has been used successfully to learn to detect object keypoints for downstream manipulation tasks \cite{dense_object_nets_learning_dense_visual_object_descriptors_by_and_for_robotic_manipulation, s3k_self_supervised_semantic_keypoints_for_robotic_manipulation_via_multi_view_consistency}, design more predictive forwards dynamics model \cite{learning_3d_dynamic_scene_representations_for_robot_manipulation, 3d_oes_viewpoint_invariant_object_factorized_environment_simulators} and improves the performance of grasping in clutter system \cite{volumetric_grasping_network_real_time_6_dof_grasp_detection_in_clutter}.  

\begin{figure}
  \centering
  \includegraphics[width=\linewidth]{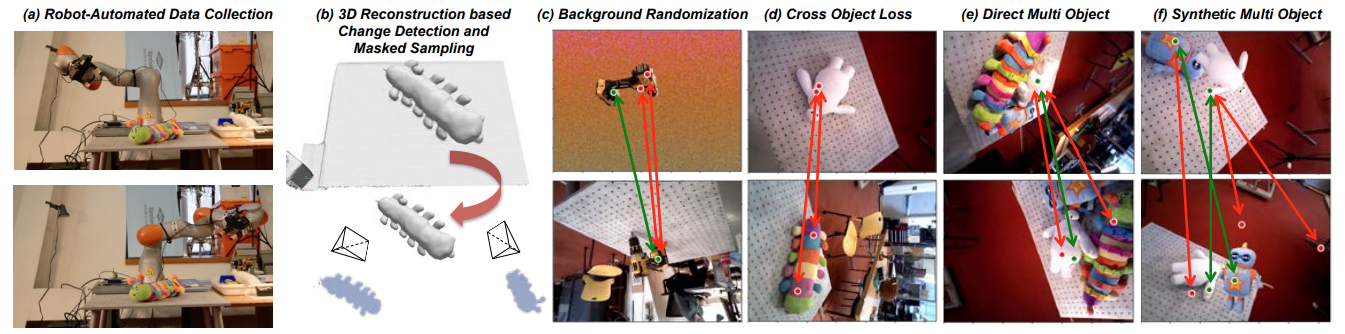}
  \caption{\textbf{Data collection and training pipeline of DenseObjectNets \cite{dense_object_nets_learning_dense_visual_object_descriptors_by_and_for_robotic_manipulation}.} (a) A robot obtains different RGBD views of an object. (b) Dense 3D recontruction and automatic object masking. (c-f) Different training strategies to improve the learned visual descriptor. Green arrows depict pixel correspondences between multiple views. Red arrows depict non-correspondences.}
  \label{fig:structure_for_learning_dense_object_net}
\end{figure}

Object keypoints are a popular representation in many computer vision tasks. \cite{dense_object_nets_learning_dense_visual_object_descriptors_by_and_for_robotic_manipulation} proposes a framework to learn to detect object keypoints without human annotation based on geometric prior and demonstrates applications of such keypoint-based object representation for manipulation tasks (\autoref{fig:structure_for_learning_dense_object_net}). To obtain a visual descriptor of an object keypoint, they first find pixel-wise correspondence between multiple views of the object. Given a static object, they move a RGBD camera mounted on a robot wrist to capture views of the object from different angles and perform dense 3D reconstruction. Two pixels from two images corresponding to two different views are a match if they correspond to the same vertex in the 3D reconstruction. Given the pixel-wise correspondence, they train a network to output per-pixel descriptor through contrastive learning. Depending on the data sampling scheme during training, they demonstrate that the keypoints and their descriptors can generalize across different instances of the same object class, across different classes of objects and are also invariant to viewpoint, object configuration and deformation. Using the learned descriptor, they demonstrate the ability to grasp an object at pre-defined locations while only requiring weak human annotation.
To do so, a human can annotate a picture of an object by clicking on the object to indicate where the robot should grasp. 
When the robot observes the object from a different view, the learned network finds a pixel location in the current view that has similar descriptor value as the pixel annotated by the human. 
The robot proceeds to grasp the object at the found pixel location in the current view, using geometric method to generate grasp proposals. 
They also demonstrate the benefit of the keypoint representation in a constrained manipulation setting, where the manipulant does not have 6 DoF in free space \cite{kpam_keypoint_affordances_for_category_level_robotic_manipulation}. \cite{s3k_self_supervised_semantic_keypoints_for_robotic_manipulation_via_multi_view_consistency} further demonstrates how to obtain such semantic keypoints with only access to RGB views and without requiring accurate depth sensing. 
Using neural network to learn pixel-wise descriptor has also seen successes when deployed to a mobile manipulation system \cite{a_mobile_manipulation_system_for_one_shot_teaching_of_complex_tasks_in_homes}, which allows human operator to teach robot primitive skills via a Virtual Reality interface. 
The robot chooses which skills to perform based on how similar the learnt visual description of the current scene is to the stored visual descriptors associated with the taught skills.

\begin{figure}
  \centering
  \includegraphics[width=0.8\linewidth]{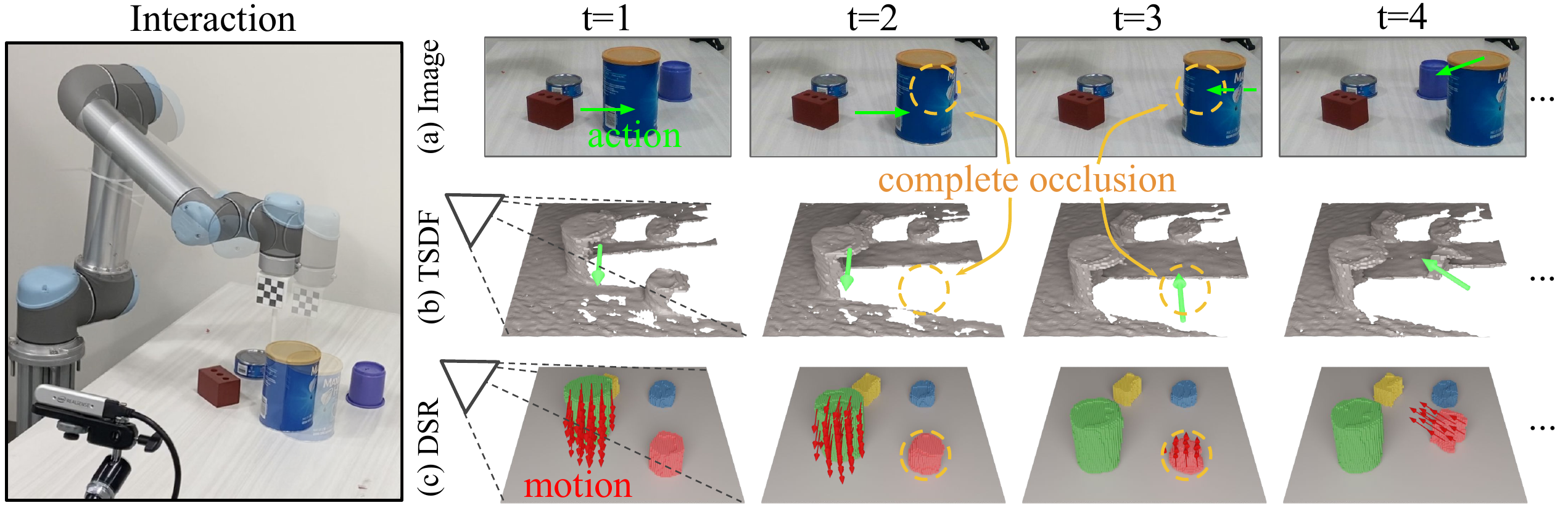}
  \caption{ \textbf{3D Dynamic Scene Representation from \cite{learning_3d_dynamic_scene_representations_for_robot_manipulation}.} (a) Color images used for illustrating the scene only (not robot-view). (b) The scene is encoded as a Truncated Sign Distance Function from depth observation. From $t=1$ to $t=3$, the blue container closer to the camera moves to the right and blocks the blue cup from view. (c) Even though the blue cup disappears from view at $t=3$, the Dynamic Scene Representation still represents the objects (ensuring object permanence), and is able to predict the motion of the blue cup at $t=4$.}
  \label{fig:structure_for_learning_dynamic_scene_representation}
\end{figure}

A common approach to apply learning to robotics problem is to train a dynamics model and then use it for planning. 
An advantage of this approach is that the networks can take raw visual observation as input.
We can also train the networks with robot interaction data, which can be easier to acquire than human annotation. 
Previous approaches have mostly modeled only the visible surfaces of objects, such as mapping RGB image to RGB image by predicting optical flow. \cite{learning_3d_dynamic_scene_representations_for_robot_manipulation} argues that such approaches perform poorly in cluttered environments, where objects frequently go out of view. They thus argue for explicit 3D structure in the representation of the learnt dynamics. They aim to achieve object permanence, amodal completeness and consistent object tracking over time. Their method infers the amodal 3D mask of each object in the scene, predicts the motion of each object, and spatially wrap the scenes using the predicted motion. 
The wrapped scene is used as input in the next step prediction to ensure objects remain in the 3D representation even though they disappear from the current view. 
\autoref{fig:structure_for_learning_dynamic_scene_representation} provides a simple example to illustrate the benefit of the proposed dynamic scene representation.

In addition to the two research direction above, 3D prior has also demonstrated benefit in grasp synthesis for cluttered scene. \cite{volumetric_grasping_network_real_time_6_dof_grasp_detection_in_clutter} argues that by including the full 3D information of the cluttered scene as input into a neural network, the network can reason about collision between the gripper and object in the scene. The network can thereby directly synthesize collision-free grasp without requiring the use of an external collision checker during grasp proposal. Their method thus only requires $10$ ms to propose grasps, which allows for more robust closed-loop grasp planning. \cite{multifingered_grasp_planning_via_inference_in_deep_neural_networks_outperforming_sampling_by_learning_differentiable_models} also proposes a voxel-based neural network to perform grasp synthesis for multi-fingered hand.

\subsection{Action parameterization}
\label{sec:structure_for_learning_act_param}

In the first waves of success in applying deep learning to robotics, many works follow the paradigm of mapping sensory information to torque directly using a neural network \cite{deep_rl_for_robotic_manipulation_with_asynchronous_off_policy_updates, learning_hand_eye_coordination_for_robotic_grasping_with_deep_learning_and_large_scale_data_collection, end_to_end_training_of_deep_visuomotor_policies}. Torque space is a relatively low-level and unstructured action space with which to send commands to the robot. Subsequent works have designed more structured action space to improve learning efficiency. 

\begin{figure}
  \centering
  \includegraphics[width=0.7\linewidth]{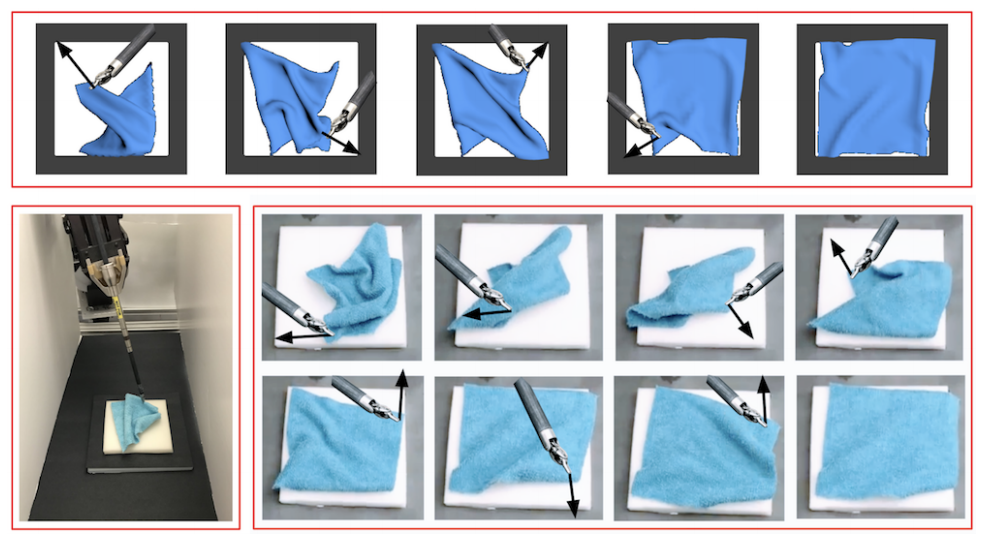}
  \caption{ \textbf{Policy execution for deformable object manipulation from \cite{deep_imitation_learning_of_sequential_fabric_smoothing_policies}.} (top row) The policy learns to flatten the cloth in simulation by imitating an expert (borrom rows) Execution on a physical da Vinci surgical robot. The action space consists of a pick location and a pull direction in the 2D plane.}
  \label{fig:structure_for_learning_action_param_pull_cloth}
\end{figure}

A major theme in designing new action parameterization is to manually design low-dimensional discrete action space instead of sending torques, a high dimensional continuous action space, to the robot. For learning how to manipulate deformable object, \cite{learning_to_manipulate_deformable_objects_without_demonstrations} designs a pick-and-place action space that encodes the conditional dependency of the optimal picking location and optimal placing location. 
They learn the placing policy conditioned on random pick points during training. 
During testing, they choose the pick points that maximize the expected value under the corresponding optimal placing locations. \cite{learning_to_manipulate_deformable_objects_without_demonstrations} trains the policy using model-free Reinforcement Learning and transfer to real robot using Domain Randomization. 
Allowing the Reinforcement Learning agents to take actions in the low-dimensional action space increases the probability the agents will perform actions that have positive rewards, thereby improving learning efficiency compared to agents operating in torque space. \cite{deep_imitation_learning_of_sequential_fabric_smoothing_policies} also tackles deformable object manipulation. 
However, they train their policy to imitate a state-based expert controller in simulation. They also design a low-dimensional pick-and-pull action space, which makes it easier for the policy to imitate the expert controller. The main limitations of these works are that the tasks considered are fairly simple, such as flattening or straightening objects (\autoref{fig:structure_for_learning_action_param_pull_cloth}). 
That being said, these tasks are still challenging due to the lack of a canonical state representation and the complex dynamics of highly deformable objects like cloth.

\begin{figure}
  \centering
  \includegraphics[width=0.5\linewidth]{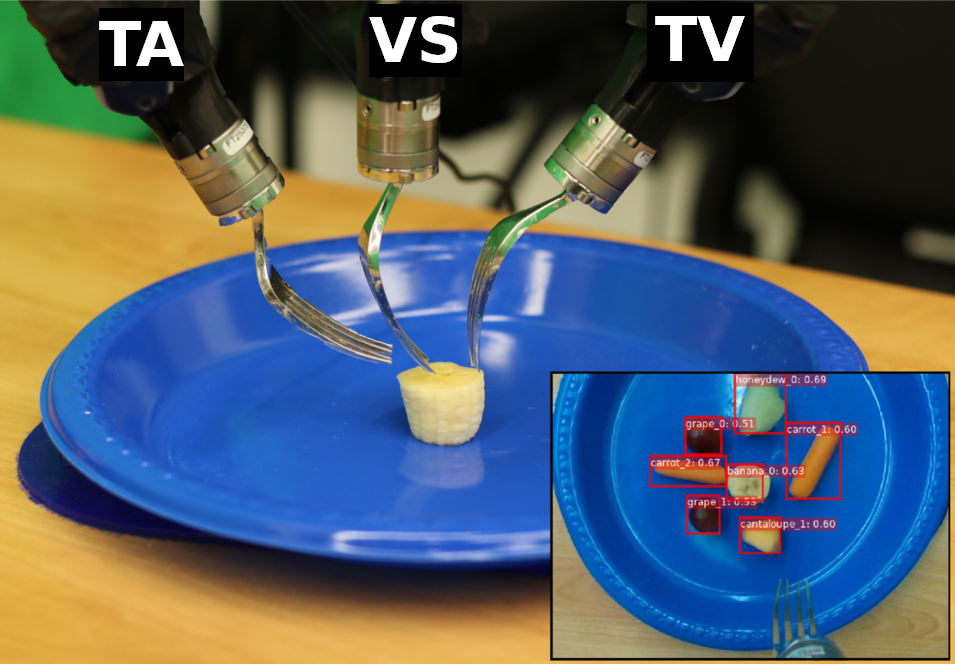}
  \caption{ \textbf{Three different fork pitch angles from \cite{learning_from_failures_in_robot_assisted_feeding_using_online_learning_to_develop_manipulation_strategies_for_bite_acquisition} .} When faced with an unknown food item, the contextual bandit algorithm explores the three different fork pitch angles (TA: tilted-angle, VS: vertical, TV: tilted-vertical) and two different roll angles (not shown) to find successful strategies to pick up a food item with the fork.}
  \label{fig:structure_for_learning_action_param_assisted_feeding}
\end{figure}

When turning the continuous action space into a low-dimensional discrete one, it is also possible to simplify the problem further by assuming the decision making problem is one-step, rather than  multi-step. For example, \cite{learning_from_failures_in_robot_assisted_feeding_using_online_learning_to_develop_manipulation_strategies_for_bite_acquisition} tackles the challenge of picking up unseen food items using a fork in an assistive feeding scenario. They design simple food picking action primitives consisting of three different pitch and two different roll angles of the fork when it approaches the food items. As such, they can apply contextual bandit algorithms with well-known regret bound to ensure convergence to a successful food picking strategy. \autoref{fig:structure_for_learning_action_param_assisted_feeding} illustrates the three different fork pitch angles. A similar simplification can be found in \cite{exploratory_grasping_asymptotically_optimal_algorithms_for_grasping_challenging_polyhedral_objects}, which tackles the task of learning to grasp challenging objects. They assume an object has a small number of stable poses and that their perception system can differentiate among these poses. For each stable pose of the object, they sample a small number of potentially successful grasp proposal from DexNet \cite{dex_net_2.0_deep_learning_to_plan_robust_grasps_with_synthetic_point_clouds_and_analytic_grasp_metrics} and let these proposals form the discrete action space associated with the stable pose. 
They thus maintain as many instantiations of bandit algorithms as there are stable poses and attempts to continuously grasp an object until a good grasp can be found. 
These methods benefit from the one-step nature of the simplified decision making problem due to having a much smaller action space to explore.

Another method to improve the exploration capability of a Reinforcement Learning agent is to augment the torque-based action space with a motion planner \cite{motion_planner_augmented_rl_for_robot_manipulation_in_obstructed_environments}. 
They train a Reinforcement Learning agent to predict the changes in joint values. If the predicted changes are small, they use low-level torque controller to achieve the desired changes. This allows the robot to perform manipulation actions requiring precision, such as those involving contacts. If the changes are large, they use a motion planner to find joint trajectory for the robot to arrive at the final desired joint position. 

A less active, though interesting research direction, is to design the action space from a control perspective. \cite{learning_variable_impedance_control_for_contact_sensitive_tasks} shows that instead of letting the neural network predicts desired torque directly, having it outputs the desired joint positions and convert the desired joint positions into torques using a variable gain PD controller leads to better performance in tasks involving hybrid motion-force control. 

\subsection{Integration of language}
\label{sec:structure_for_learning_language}

The advances of deep learning in natural language understanding have also motivated the integration of these techniques into robotics. This has enabled the use of language-based semantics to improve the generalization of grasping systems \cite{same_object_different_grasps_data_and_semantic_knowledge_for_task_oriented_grasping} and specifies tasks to the robots in a more intuitive manner \cite{concept2robot_learning_manipulation_concepts_from_instructions_and_human_demonstrations, robot_object_retrieval_with_contextual_natural_language_queries}. 

\begin{figure}
  \centering
  \includegraphics[width=\linewidth]{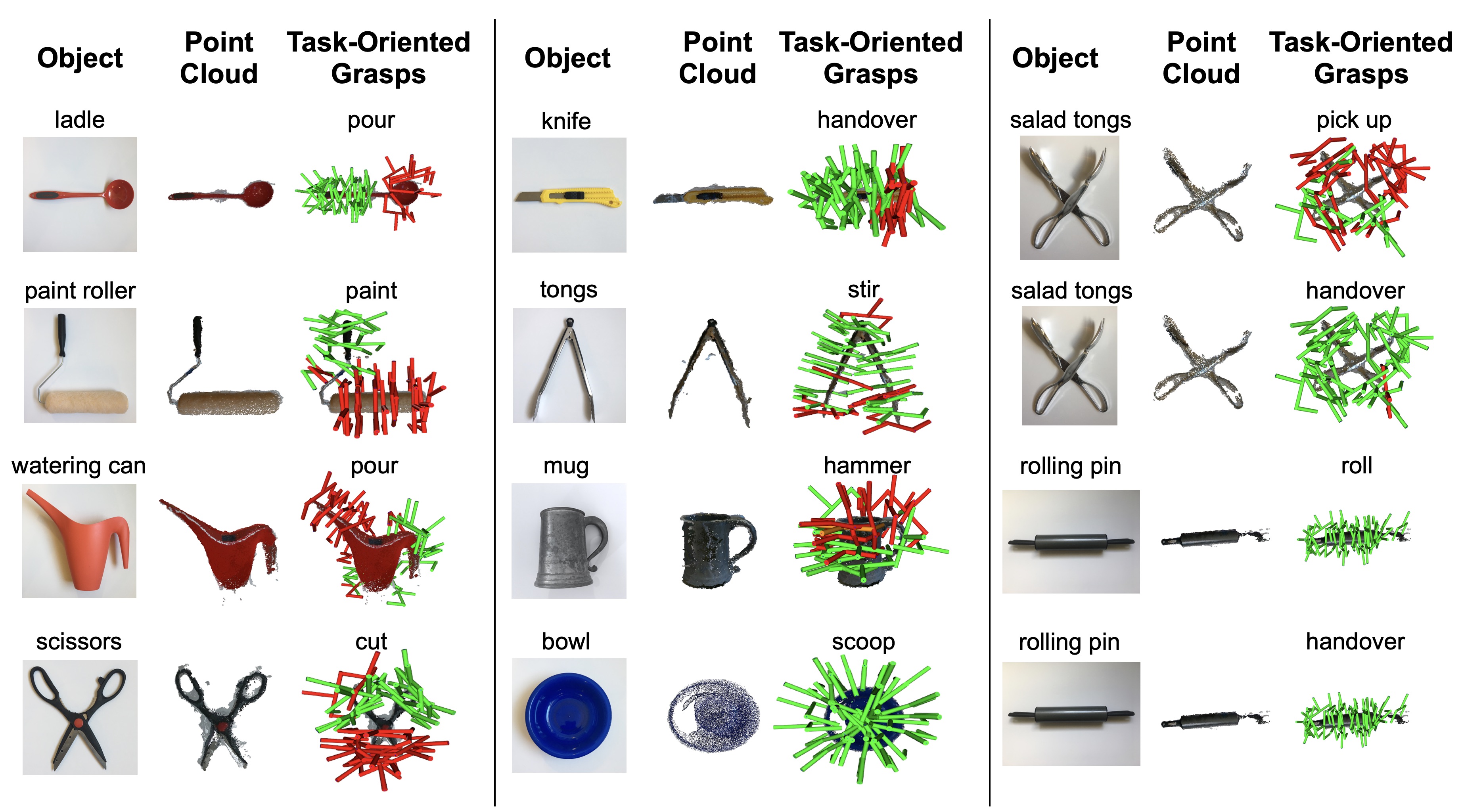}
  \caption{ \textbf{Examples of human-annotated task-oriented grasps from \cite{same_object_different_grasps_data_and_semantic_knowledge_for_task_oriented_grasping}.} For each object, the authors provide RGB images, point cloud and stable grasps which were chosen by human annotators to satisfy a particular task, such as pick up or handover.}
  \label{fig:structure_for_learning_language_same_object_different_grasp}
\end{figure}

\begin{figure}
  \centering
  \includegraphics[width=0.5\linewidth]{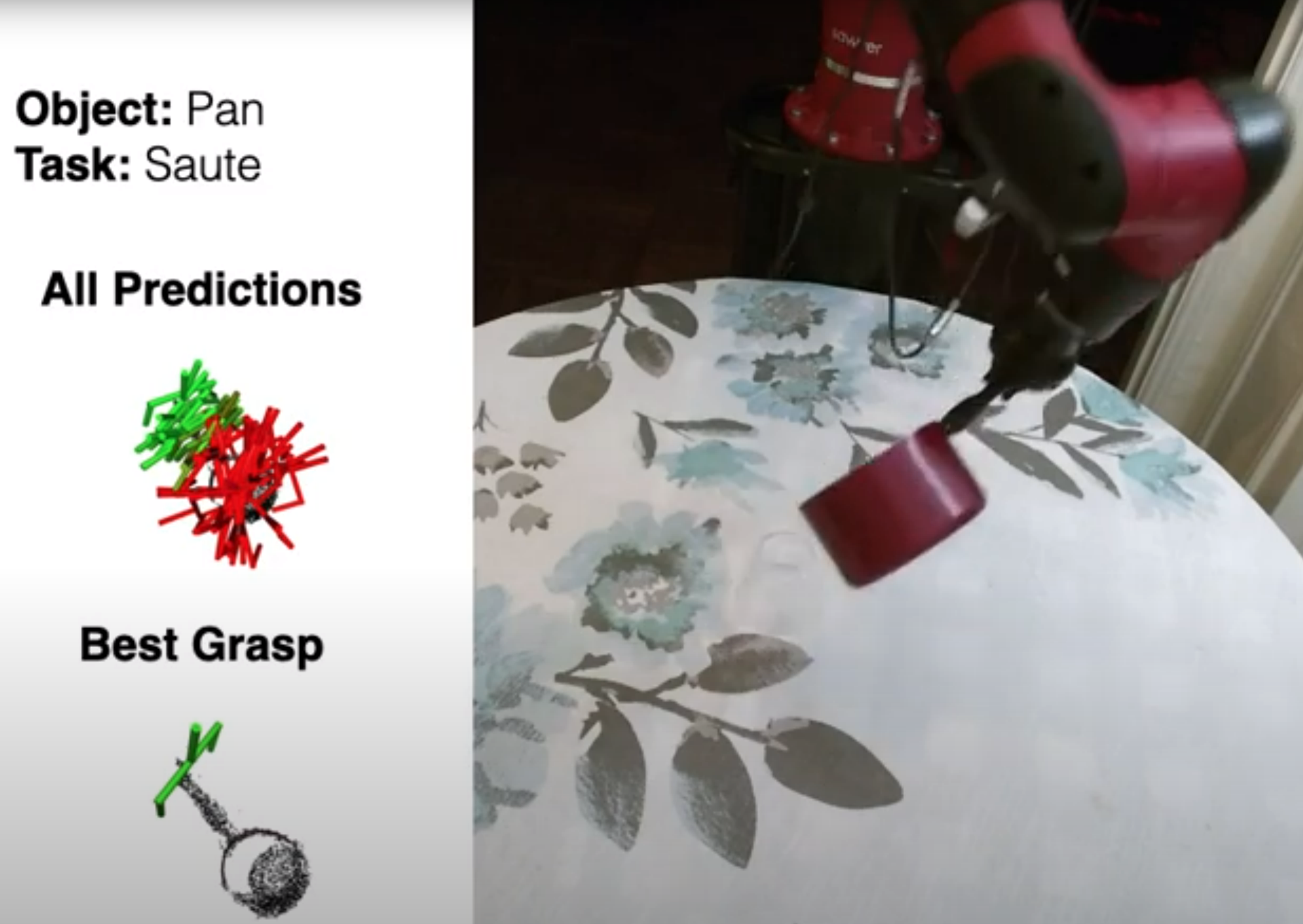}
  \caption{ \textbf{Failed grasp proposal from \cite{same_object_different_grasps_data_and_semantic_knowledge_for_task_oriented_grasping}.} Since their method does not take into account post-grasp object-gripper interaction, the configuration of the grasped object would not allow the manipulator to perform the task in this case. The task is ``saute", which would require the pot bottom to be roughly parallel to the ground. However, the top-down grasp instead flips the pot such that the pot's bottom is perpendicular to the ground instead.}
  \label{fig:structure_for_learning_language_same_object_different_grasp_failure}
\end{figure}

\cite{same_object_different_grasps_data_and_semantic_knowledge_for_task_oriented_grasping} proposes a new dataset and algorithm to generate task-oriented grasp. The dataset contains RGB images and pointclouds of $191$ different objects from $75$ different object categories. There are also $56$ different task labels, such as pick up and handover. 
For each object, human annotators label which tasks a stable grasp for the object will allow the manipulator to perform out of the $56$ different task labels. 
\autoref{fig:structure_for_learning_language_same_object_different_grasp} provides examples of the objects in the dataset and their corresponding task-oriented grasp labels. 
To predict the task-oriented grasp for an object, they encode the object categories and the tasks they afford in a knowledge graph. 
The feature for each node in the graph is the word embedding of the corresponding word at the node. 
They then perform message passing to predict the task-oriented grasp and train the network end-to-end. 
They demonstrate that the incorporation of the semantic knowledge of the object category and the tasks in terms of their word embedding and expressing relationships between the words in a knowledge graph lead to better generalization to unseen objects. A limitation of their approach is that they do not consider the post-grasp object-gripper interaction, which can lead to false positive in the generated task-oriented grasp. \autoref{fig:structure_for_learning_language_same_object_different_grasp_failure} provides an example for such a false positive.

\begin{figure}
  \centering
  \includegraphics[width=0.5\linewidth]{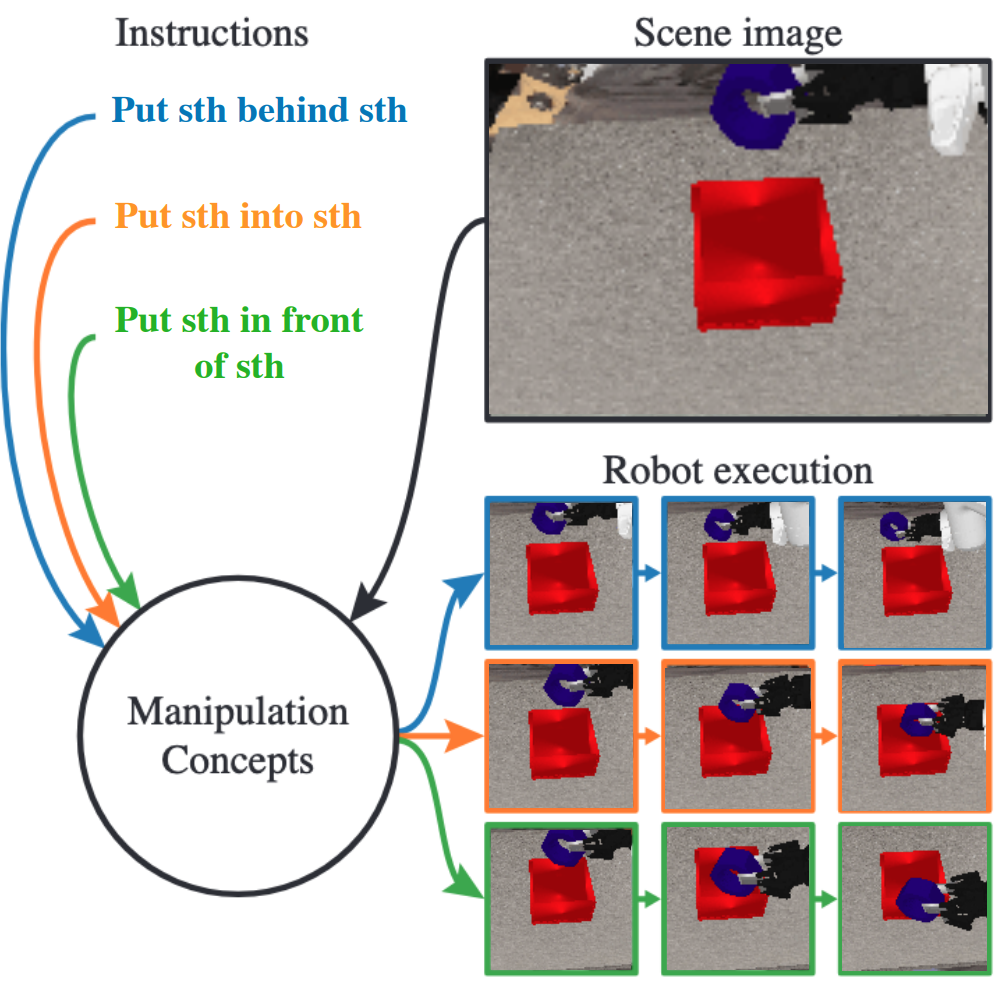}
  \caption{ \textbf{Examples of language instructions from \cite{concept2robot_learning_manipulation_concepts_from_instructions_and_human_demonstrations} .} Given an image of the current scene and a language instruction, the controller should generate a robot trajectory that accomplishes the task specified in the instruction. The colored arrow indicates the different language instructions and the corresponding motion trajectories.}
  \label{fig:structure_for_learning_language_concept2robot}
\end{figure}

We can also use language to specify tasks that the robot should perform \cite{concept2robot_learning_manipulation_concepts_from_instructions_and_human_demonstrations}. Given an image of the current scene and language instructions, the policy from \cite{concept2robot_learning_manipulation_concepts_from_instructions_and_human_demonstrations} generates a motion trajectory that accomplishes the tasks specified in the instructions. \autoref{fig:structure_for_learning_language_concept2robot} provides a few examples of the language instructions considered. Their method consists of two phases. First, multiple single-task policies are trained independently through RL. Second, they train a multi-task policy to combine the single-task policies such that different single-task policies are executed based on the input language instruction. To train the single-task policy, they use a video-based action classifier to judge how well the robot visually appears to perform the tasks. A potential limitation of this approach is that the video-based action classifier is trained from human activity dataset. As such, it might not be suitable to judge how well the robot is performing a task for more complex tasks, since a robot might perform tasks differentially from human due to their different body morphologies. 

\cite{robot_object_retrieval_with_contextual_natural_language_queries} also considers using language to specify tasks for a robot, but the tasks they consider are more constrained. Given a language command containing a verb, such as ``hand me something to \textit{cut}" and RGB images of potential candidates objects, the robot should pick up the object that will best allow the human to perform the task with. Their method can generalize to unseen object classes and unknown nouns in the language commands. 

\section{Using learning to improve model-based methods}
\label{sec:learning_to_improve_model_based_methods}

While the methods presented in the previous sections predominantly use additional structure to improve the learning efficiency, researchers have also explored using learning techniques to improve the performance of more classical model-based methods.

\begin{figure}
  \centering
  \includegraphics[width=0.6\linewidth]{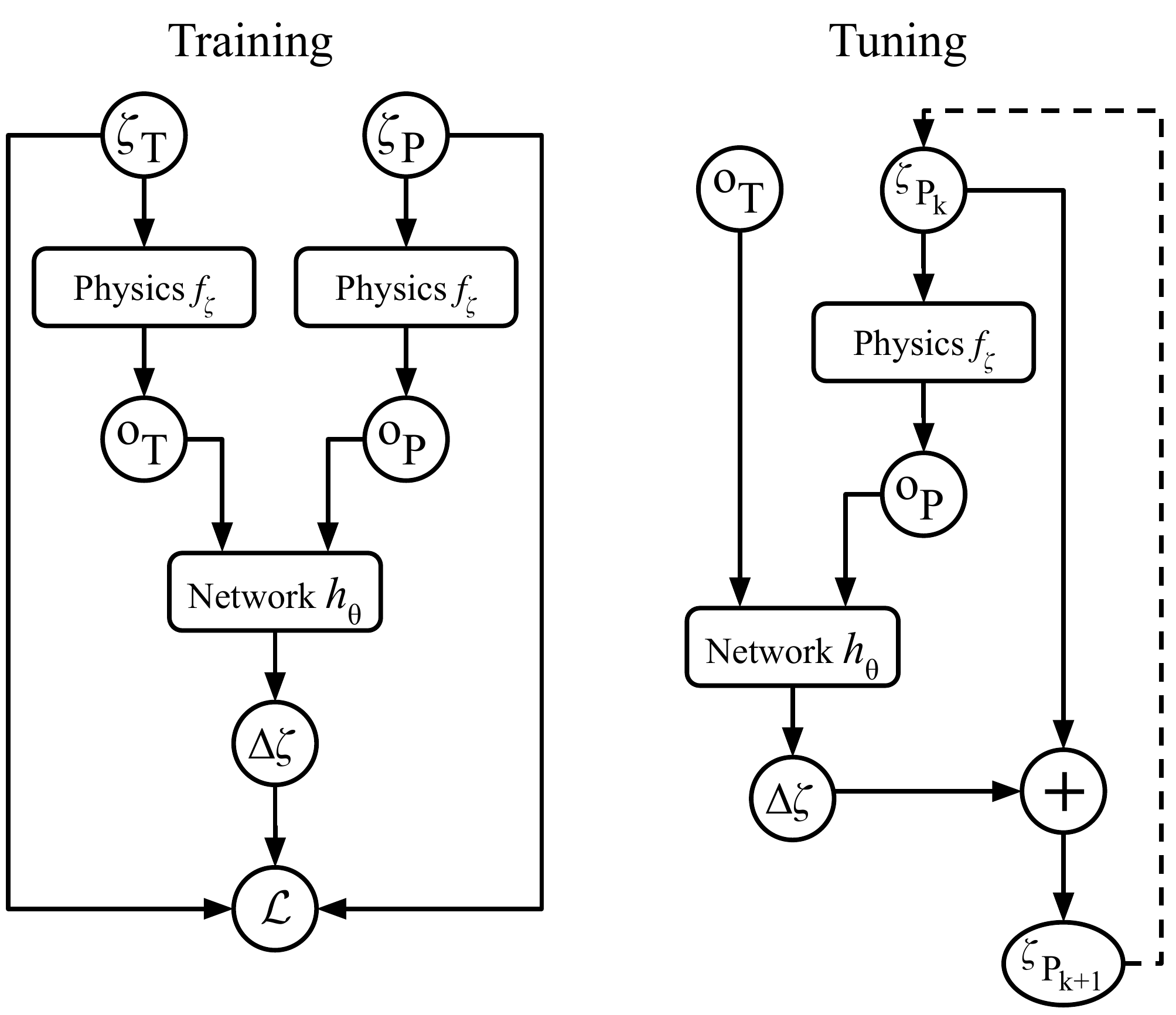}
  \caption{ \textbf{Training and inference procedure of TuneNet \cite{tunenet_one_shot_residual_tuning_for_system_identification_and_sim_to_real_robot_task_transfer}.} During training, given observation $o_T$ from the target model and observation from the current model $o_P$, they train the network $h$ with parameters $\theta$ to output a residual $\Delta \zeta $ such that adding the residual to the current physical simulation parameters $\zeta_P + \Delta \zeta$ reduces the differences between the observations from the target and current model. During testing, $h_\theta$ takes as input the target and current observations and outputs the residual $\Delta \zeta$. The new estimate is the sum of the residual and the current estimate.}
  \label{fig:learning_for_physics_tunenet}
\end{figure}

\cite{tunenet_one_shot_residual_tuning_for_system_identification_and_sim_to_real_robot_task_transfer} modifies the parameters of a physical simulator such that the simulation better matches a target physical model. There are two main insights offered in their paper. First, estimating the residual between the current simulation parameters and the target simulation parameters is more effective than predicting the target simulation parameters directly. They argue this is because small differences are more frequently observed in the training dataset. Second, they generate their training data using only simulation. However, their neural network trained with such synthetic data can generalize to real world scenarios. The training and inference procedures are illustrated in \autoref{fig:learning_for_physics_tunenet}. They demonstrated their method in estimating the coefficient of restitution of a bouncing ball. The observation in this case is the position of the ball. The main limitation of this work is the need to carefully measure the observation from the target model, which might be challenging to obtain for more complex object interactions. They also require the arrangement of the objects in simulation to roughly match those in the target model, to impose constraints on the estimation problem and allow estimating the residuals with a small number of data coming from the target model.

In addition to improving the model, we can also use learning to improve the controller.
To improve Feedback Linearization control, \cite{an_online_learning_procedure_for_feedback_linearization_control_without_torque_measurements} performs online learning to generate torque compensation to account for unmodeled dynamics. 
As discussed in \autoref{sec:background_control}, given a perfect model of the robot, we can cancel out the non-linear dynamics in the computed torque control law, leading to a stable linear error dynamics. 
The model of the robots is often computed offline before robot deployment. 
However, during execution, the model might change, for example, when the robot picks up a box. 
In such case, the large change in the structural property of the robot inhibits the ability of the computed torque control law to accurately track the desired robot motion. \cite{an_online_learning_procedure_for_feedback_linearization_control_without_torque_measurements} interprets the difference in the desired and achieved motion as if Feedback Linearization was accurate, but the robot was driven by an unknown acceleration reference. They thus estimate the unknown acceleration reference with the Controllability Gramian, which allows for acceleration estimation without resorting to numerical differentiation. The estimated acceleration forms a dataset with which they train a Gaussian Process Regression model to generate the torque compensation for the unmodeled dynamics. They emphasize that their main difference compared to previous work is that they do not use torque measurement, which is often noisier than joint position and velocity measurement. It would be interesting to compare their approach against a dynamic gain PID controller in terms of the effectiveness in tracking desired motion in the presence of large unmodeled dynamics.

\section{Data collection methods}
\label{sec:data_collection_methods}

Learning methods require a high amount of interactions with their environment to exhibit interesting behaviors. In this section, we discuss how successful applications of learning methods have instrumented either the environment \cite{deep_dynamics_models_for_learning_dexterous_manipulation, tossingbot_learning_to_throw_arbitrary_objects_with_residual_physics, form2fit_learning_shape_priors_for_generalizable_assembly_from_disassembly} or the robots \cite{visual_imitation_made_easy, grasping_in_the_wildlearning_6dof_closed_loop_grasping_from_low_cost_demonstrations, dexpilot_vision_based_teleoperation_of_dexterous_robotic_hand_arm_system, s4g_amodal_single_view_single_shot_se3_grasp_detection_in_cluttered_scenes} to automate the data collection process and increase the usefulness of the collected data.

We split the section into autonomous data collection (\autoref{sec:data_collection_auto}) and human-assisted data collection (\autoref{sec:data_collection_human_assist}).

In addition to the works discussed below, \cite{learning_latent_plans_from_play, relay_policy_learning_solving_long_horizon_tasks_via_imitation_and_rl} considers how to better make use of structure present in existing datasets.

\begin{figure}
  \centering
  \includegraphics[width=0.8\linewidth]{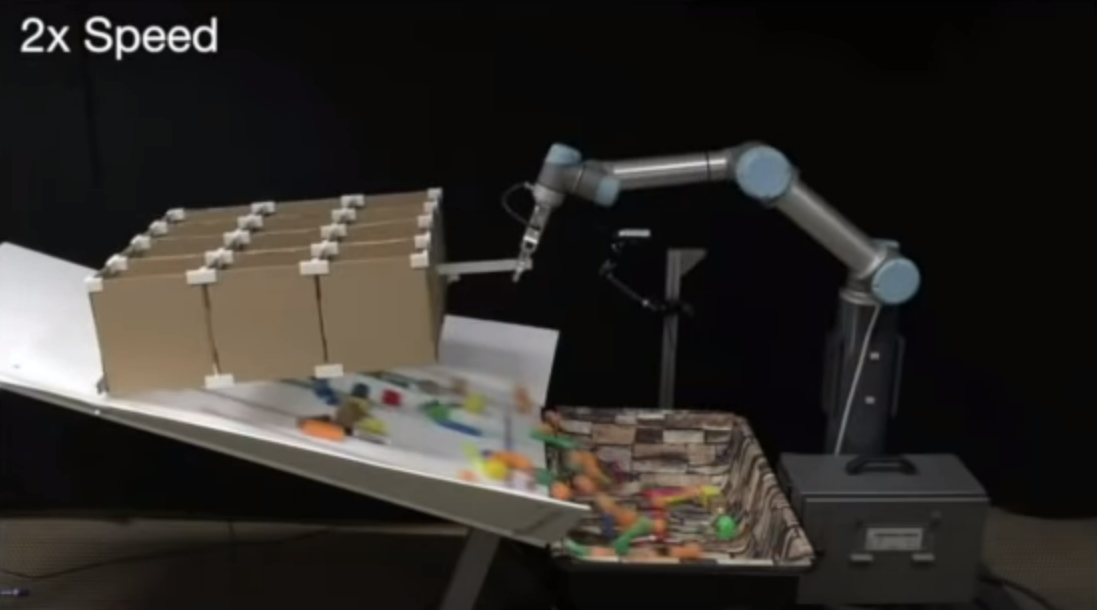}
  \caption{ \textbf{Environment reset mechanism from \cite{tossingbot_learning_to_throw_arbitrary_objects_with_residual_physics}.} The robot attempts to throw objects into the cardboard bins. After most objects have been thrown, the robot lifts the cardboard bins. The objects in the bins flow back into the tray in front of the robot due to gravity. With such reset mechanism, the robot can attempt the throws again without requiring human assistance.}
  \label{fig:data_collection_tossing_bot}
\end{figure}

\subsection{Autonomous data collection}
\label{sec:data_collection_auto}

Environment reset is a critical mechanism that allows robots to autonomously collect data. To enable environment reset, we need to instrument the environment of the robot such that we can reset the state of the environment back to an initial configuration. If the robot's attempt at performing the tasks fails, it invokes the environment reset mechanism to allow it to try to perform the tasks again. This removes the need for human intervention, which would represent a bottleneck in the data collection pipeline. 

A recent successful demonstration of learning for manipulation is \cite{tossingbot_learning_to_throw_arbitrary_objects_with_residual_physics}. They propose a system that allows robots to throw arbitrary objects into bins. This is a challenging task because the ballistic trajectories and the final locations of the thrown objects are a complex function of the throwing forces, gripper-object contact locations and object's inertial and material properties. The environment reset mechanism in \cite{tossingbot_learning_to_throw_arbitrary_objects_with_residual_physics} allows their robot to repeatedly try to throw objects into randomly chosen bins. Even if the thrown objects do not end up in the correct bins, it would arrive at the neighboring bins, which allow for effective environment reset. \autoref{fig:data_collection_tossing_bot} illustrates the environment set-up that allows for reset.

\begin{figure}
  \centering
  \includegraphics[width=0.45\linewidth]{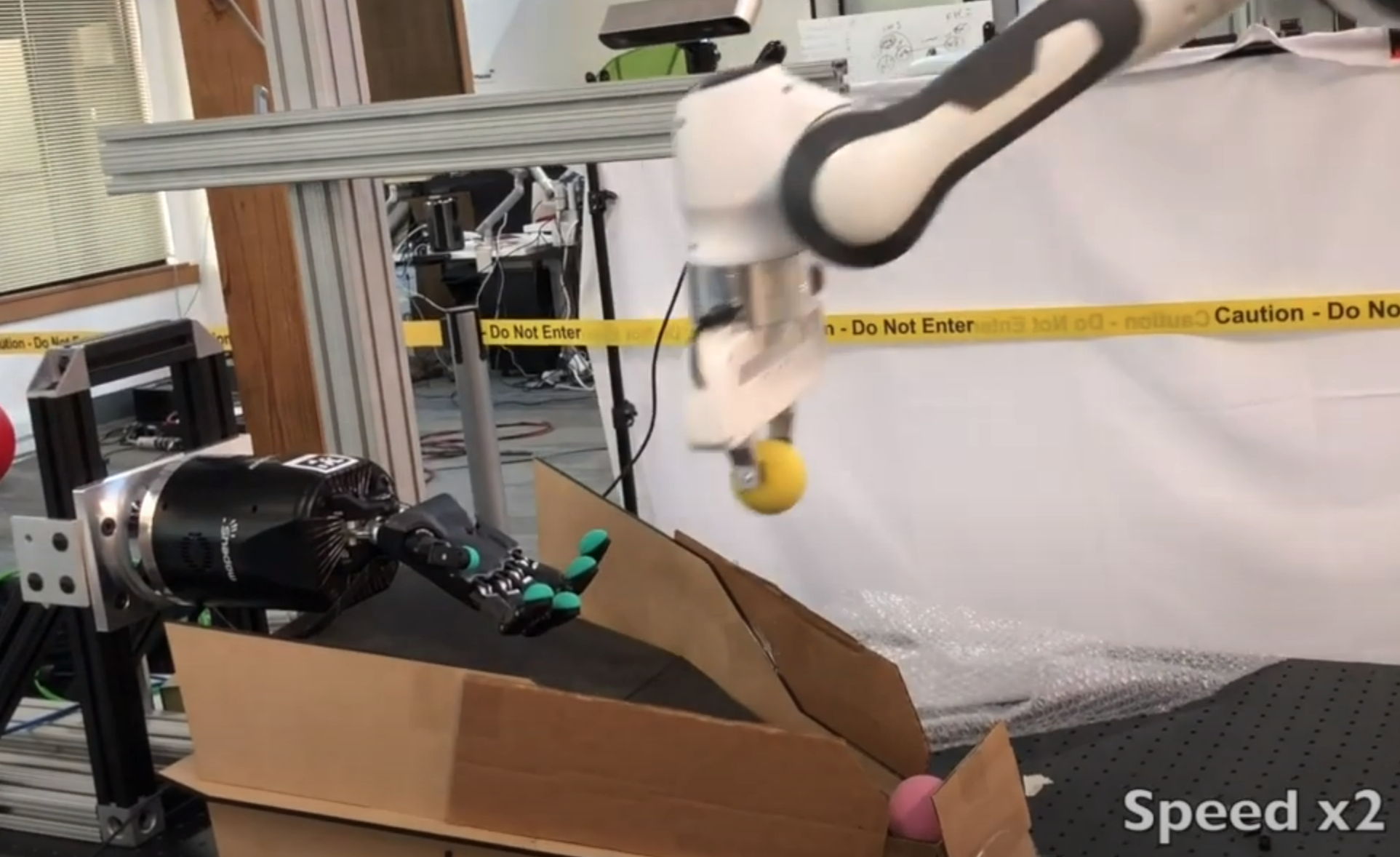}
  \caption{ \textbf{Environment reset mechanism from \cite{deep_dynamics_models_for_learning_dexterous_manipulation}.} The 24-DoF anthropomorphic hand learns how to rotate balls on its palm without dropping them. Note that the hand faces upwards, therefore making use of gravity to ensure the balls are likely to stay inside the palm. If the hand drops the ball during the course of learning, the white robot picks up the ball and places them inside the palm of the hand. This allows the hand to continuously learn without human intervention.}
  \label{fig:data_collection_baoding_ball}
\end{figure}

Learning has also been successfully applied to tasks requiring repeatedly making and breaking contacts, which are hard to model and estimate accurately \cite{deep_dynamics_models_for_learning_dexterous_manipulation}. The paper teaches a robot with 24-DoF anthropomorphic hand to perform complex tasks, such as moving a pencil to write or rotating baoding balls in-hand. Their method consists of a learned predictive model and a trajectory optimization method to generate actions. To enable the hand to continuously learn how to rotate the balls without dropping them, they have another robot arm which picks up the ball and places them in the palm of the multi-finger hand. \autoref{fig:data_collection_baoding_ball} illustrates this mechanism.

Other notable successes include \cite{form2fit_learning_shape_priors_for_generalizable_assembly_from_disassembly}, which learns to perform the tasks of assembly. This involves picking up objects and placing them into a specific location inside a container with tight clearance. They have two main mechanisms to reset the environment. They restrict their methods to planar setting, which allows for the design of simple motion primitives that can pick and place object. More importantly, they generate the training data by disassembly, that is, taking apart already assembled parts. This was helpful because disassembly is much easier compared to assembly for the objects and tasks they consider. Performing disassembly therefore allows them to back-play the trajectories generated from the disassembly process and treat them as successful assembly trajectories. 

Clever use of easily controlled mechanism can also generate data for object segmentation \cite{self_supervised_object_in_gripper_segmentation_from_robotic_motions}. The main limitation of these methods is the need to design task-specific environment reset mechanism. This limits their applicability and increases the amount of background engineering required to teach robot new skills.

Allowing the robots to autonomously collect data in simulations and transferring the learned policies to real world scenarios are also widely used \cite{murali20206dof, grasping_unknown_objects_by_coupling_deep_rl_generative_adversarial_networks_and_visual_servoing, learning_to_assemble_estimating_6d_poses_for_robotic_object_object_manipulation, self_supervised_sim_to_real_adaptation_for_visual_robotic_manipulation, learning_object_manipulation_skills_via_approximate_state_estimation_from_real_videos, danielczuk2020xray, learning_to_manipulate_object_collections_using_grounded_state_representations, hoque2020visuospatial}. These works highlight the benefits of having accurate approximate physical models, which allows for simulation.

\subsection{Human-assisted data collection}
\label{sec:data_collection_human_assist}

Providing low-friction methods for human to perform demonstration is another interesting line of research.

\begin{figure}
  \centering
  \includegraphics[width=0.6\linewidth]{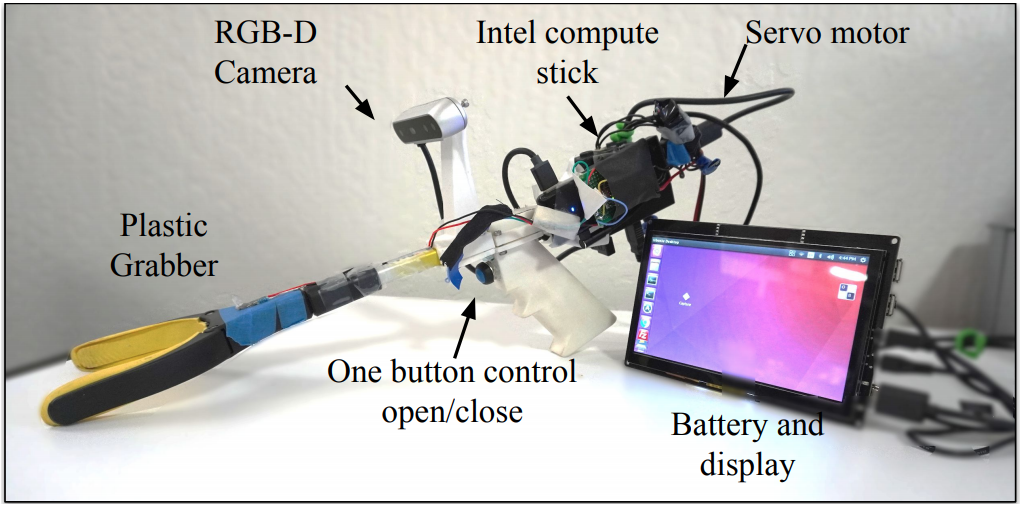}
  \caption{ \textbf{Gripper design from \cite{grasping_in_the_wildlearning_6dof_closed_loop_grasping_from_low_cost_demonstrations}.} They use simple gripper to collect grasping demonstration in more realistic scenarios compared to prior works.}
  \label{fig:data_collection_grasping_in_the_wild}
\end{figure}

\cite{visual_imitation_made_easy, grasping_in_the_wildlearning_6dof_closed_loop_grasping_from_low_cost_demonstrations} use simple, cheap and easily controlled grippers to collect demonstration for grasping. There are two main benefits. First, they can theoretically collect demonstrations on realistic objects and in more unconstrained scenarios compared to allowing the robots to collect data autonomously. When controlled by human, the chances that the robots will damage itself or its environments are significantly reduced. Since learning methods have not quite yet generalize outside the training distribution, the ability to collect data  in more unconstrained setting means that the learned policies might perform better in real-world deployment scenarios. Second, the control interface is intuitive and the grippers are cheap, thereby allowing them to scale data collection. \autoref{fig:data_collection_grasping_in_the_wild} illustrates the gripper design from \cite{grasping_in_the_wildlearning_6dof_closed_loop_grasping_from_low_cost_demonstrations}.

\begin{figure}
  \centering
  \includegraphics[width=0.6\linewidth]{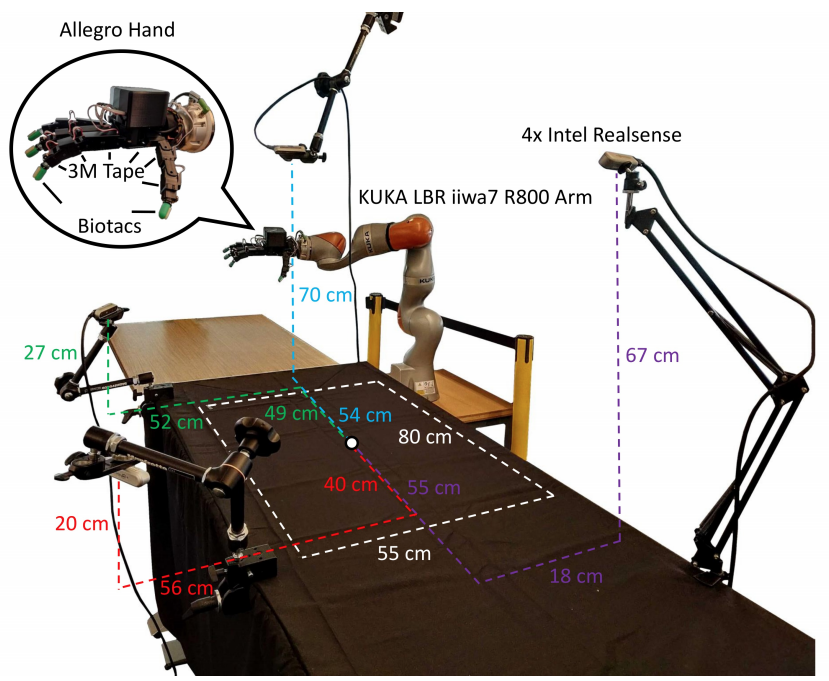}
  \caption{ \textbf{Hardware set-up from \cite{grasping_in_the_wildlearning_6dof_closed_loop_grasping_from_low_cost_demonstrations}.} They use four depth cameras to track human poses and the values of the finger joints. They perform kinematic retargeting to control a 23-DoF Allegro hand.}
  \label{fig:data_collection_dexpilot}
\end{figure}

On the other end of the end-effector spectrum, \cite{dexpilot_vision_based_teleoperation_of_dexterous_robotic_hand_arm_system} proposes a system for human to tele-operate complex multi-finger hands using only vision sensors. \autoref{fig:data_collection_dexpilot} illustrates their hardware set-up. Because their system controls a hand with many degrees of freedom, they require significantly more complex hardware system and algorithms compared to the simple gripper line of works. They use four depth cameras and a model-based tracker \cite{dart_dense_articulated_real_time_tracking} to track human hand pose and the value of the human finger joints. They then perform kinematic retargeting to map the detected human hand poses and joint angles to the poses and angles of the robot hand. Through careful tuning, they demonstrate impressive task completion such as retrieving a tea bag from a drawer.

Recent interesting works on human-in-the-loop learning also include \cite{physics_based_dexterous_manipulations_with_estimated_hand_poses_and_residual_rl, faster_confined_space_manufacturing_teleoperation_through_dynamic_autonomy_with_task_dynamics_imitation_learning}. 

\section{New datasets and benchmarks}
\label{sec:new_datasets}

It would be a travesty to discussing learning without discussing new datasets and benchmarks.

\begin{figure}
  \centering
  \includegraphics[width=0.8\linewidth]{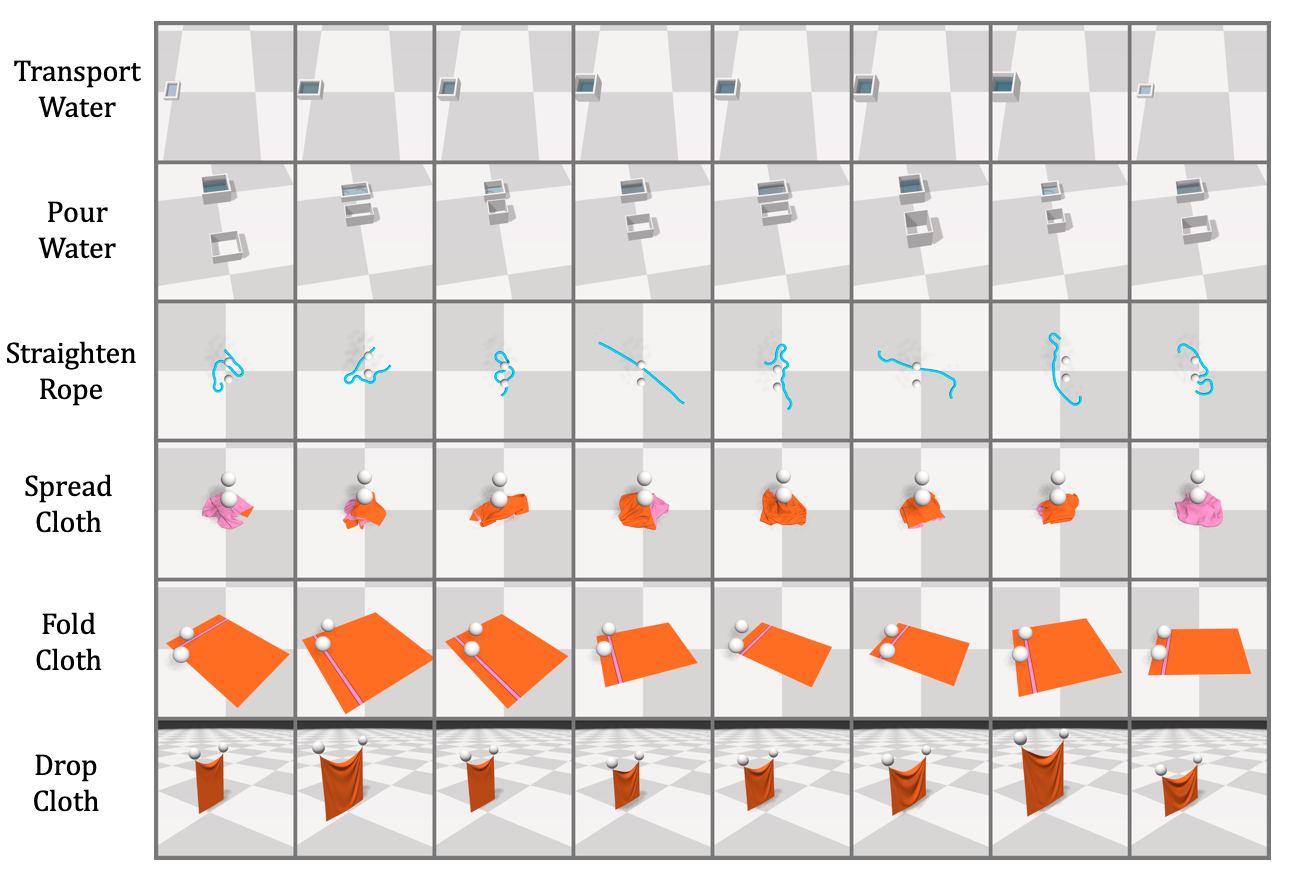}
  \caption{ \textbf{Examples of the tasks in the deformable object manipulation benchmark from  \cite{softgym_benchmarking_deep_rl_for_deformable_object_manipulation}.} They model the gripper as a picker, a spherical object that can move freely in space. The deformable objects are modeled as collection of particles. Once the picker comes close to a particle and becomes activated, the particle follows the motion of the picker.}
  \label{fig:new_dataset_softgym}
\end{figure}

\cite{softgym_benchmarking_deep_rl_for_deformable_object_manipulation} proposes a simulated benchmark for deformable object manipulation (\autoref{fig:new_dataset_softgym}). This paper fills an important missing gap in that previous simulated benchmark either provides direct access to low-dimensional state of the environment or only involves manipulating rigid bodies. The paper demonstrates the sub-optimality of current Reinforcement Learning algorithms when the input are images and the agent needs to manipulate objects with high intrinsic dimension such as cloth. They also demonstrate the realism of the simulated dynamics by comparing the result of cloth manipulation in simulation against the result of manipulating a real piece of cloth. 

\cite{robonet_large_scale_multi_robot_learning} provides a large dataset of real world robot interactions. It consists of more than $160,000$ trajectories with video and action sequences of $7$ robots interacting with hundreds of objects. To script the controller for the robots to interact with the objects, they use simple grasping primitives with added Gaussian noise. Even though the dataset consists of more than $15$ million frames, which is roughly equal in size to ImageNet, it has two main limitations. First, the dataset is still collected in a constrained lab setting, which means any models trained from the dataset are unlikely to generalize immediately to more practical settings. Second, the tasks considered are still quite simple because the scripted policies to generate data only consist of grasping primitives. I believe that methods which allow for easier data collection in unconstrained environments, as discussed in \autoref{sec:data_collection_human_assist}, will be more impactful in the future.

\section{Learning for tactile sensing}
\label{sec:learning_for_tactile}

A major advantage of learning systems is their ability to perceive and learn useful representation for high-dimensional sensory input, as evident by their successes in tasks where the input consists of images. Researchers have therefore applied learning to perceive tactile signal, which is also high-dimensional and hard to model accurately. Tactile signal provides direct sensory access to information that would otherwise be non-trivial to obtain through vision sensing alone, such as the magnitude of forces applied on objects or the objects' material properties \cite{swingbot_learning_physical_features_from_in_hand_tactile_exploration_for_dynamic_swing_up_manipulation}. Tactile information can also supplement visual information during occlusions of the object, which frequently occur in robotic manipulation \cite{tactile_localization_from_the_first_touch}.

\begin{figure}
  \centering
  \includegraphics[width=0.85\linewidth]{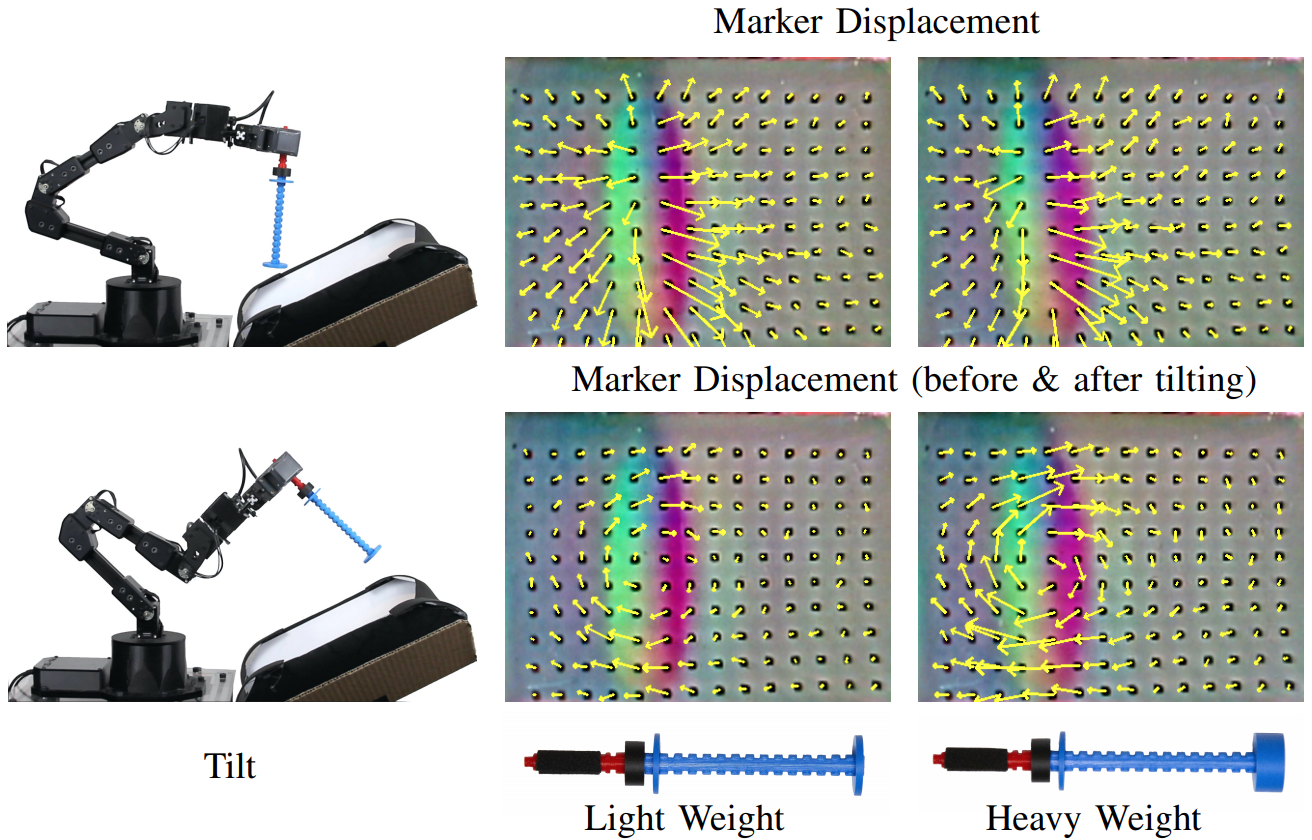}
  \caption{ \textbf{The tactile feedback from \cite{swingbot_learning_physical_features_from_in_hand_tactile_exploration_for_dynamic_swing_up_manipulation}.} Given different object properties, such as mass, the same action by the robot, tilt in this case, leads to different tactile signal profile. They use the tactile signal to learn a low-dimensional embedding of the object and use the embedding to swing objects to a chosen angle.}
  \label{fig:tactile_swingbot}
\end{figure}

In \cite{swingbot_learning_physical_features_from_in_hand_tactile_exploration_for_dynamic_swing_up_manipulation}, the robot learns a low-dimensional embedding of the physical features of a held object, such as frictional properties, from dense tactile feedback. They use the vision-based GelSight sensor  \cite{gelsight_high_resolution_robot_tactile_sensors_for_estimating_geometry_and_force}, which provides high resolution information about the contact surface between the object and the finger. They also add markers along the sensing surface to obtain the tangential displacement of objects. \autoref{fig:tactile_swingbot} provides an illustration of the tactile signal in the form of a vector field. They design two action primitives, tilting and shaking, to obtain tactile signal which can disambiguate the properties of the objects, including friction, center of mass, mass, and moment of inertia. They learn a low-dimensional embedding which fuses the data generated from the two actions. The embedding can be disentangled such that each embedding dimension corresponds to one physically meaningful property. Given the embedding, they learn a forward dynamics model which allows them to optimize for actions that swing a held object to a chosen angle to within $17\degree$ error. While swing-up is not a particular practical robotics task, the paper demonstrates the benefits of the tactile sensor and the ability to interpret the resulting sensory information with learning methods.

\begin{figure}
  \centering
  \includegraphics[width=\linewidth]{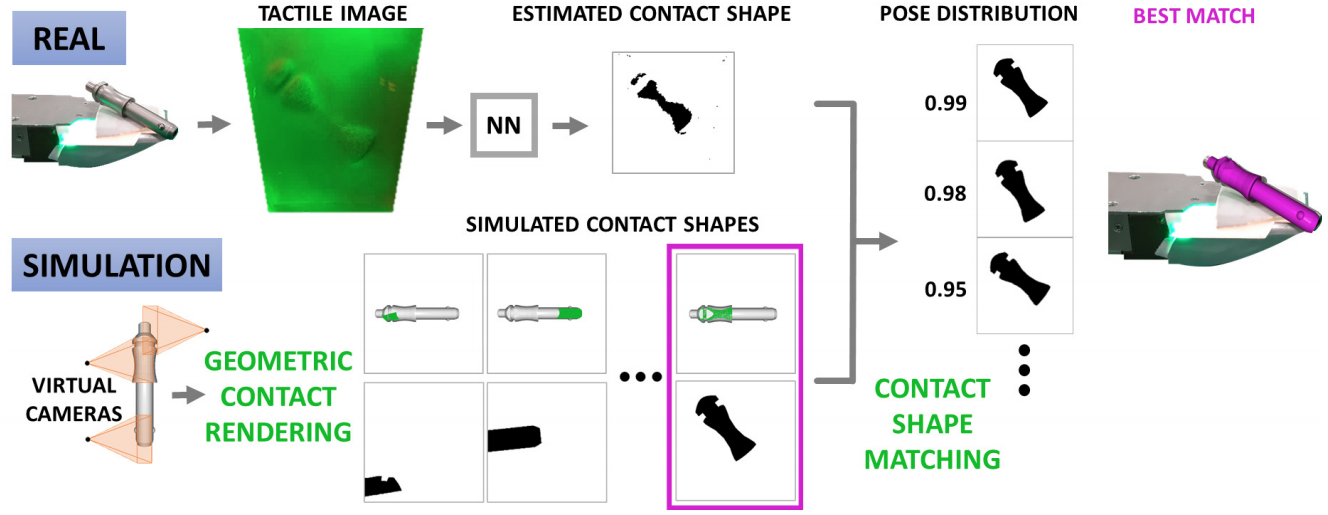}
  \caption{ \textbf{Overview of the tactile-based object pose localization system from \cite{tactile_localization_from_the_first_touch}.} (bottom row) In simulation, they render a set of object contact shape (in black) from a set of possible contact between the object and the tactile sensor (in green). (top row) Given a real object, they learn a neural network to map the tactile sensory information to the contact shape. They match this estimated contact shape against the contact shape generated in simulation to infer the most likely object pose.}
  \label{fig:tactile_object_localization_first_touch}
\end{figure}

In addition to inferring the mechanical properties of object, tactile signal can also be used to infer object's pose. \autoref{fig:tactile_object_localization_first_touch} illustrates the system from \cite{tactile_localization_from_the_first_touch} to estimate the object poses of an object from a single interaction between the tactile sensor and the object. They match the estimated contact shape against a pre-determined set of contact shapes to determine the most likely object pose that generates the estimated contact shape. To ensure efficiency, they assume access to the object 3D model and develop object-specific model. They argue these are realistic assumptions in industrial setting.

Other recent works which use learning to process high-dimensional tactile data include \cite{stable_in_grasp_manipulation_with_a_low_cost_robot_hand_by_using_3_axis_tactile_sensors_with_a_cnn} (using high frequency tactile information to achieve stable in-grasp manipulation for low-cost multi-fingered hands), \cite{towards_learning_to_detect_and_predict_contact_events_on_vision_based_tactile_sensors} (predicting contact events such as slipping from tactile information to improve robustness of closed-loop grasping) and \cite{mat_multi_fingered_adaptive_tactile_grasping_via_deep_rl} (proposing a novel curriculum of action motion magnitude which makes learning more tractable).

\section{Conclusion}

The applications of machine learning to robotic manipulation have enabled new capabilities and opened up interesting research venues. That being said, the successes of machine learning in robotic manipulation tasks have in some ways seemed more limited than the successes in computer vision or natural language understanding. We can venture guesses why this is the case. First, robotic manipulation research is significantly more open-ended and the community does not have a set of mutually agreed upon ``north star" problems. For this reason and the need to run experiments on physical robots, researchers often do not compare their algorithms against previous works. The lack of reproducibility and a common set of challenges mean researchers often demonstrate their algorithms in simple tasks, such as block pushing and stacking. Thus, despite an outpouring of research, it is unclear whether the community as a whole is making progress on the challenges of robotic manipulation. Second, it is challenging to collect realistic large-scale datasets to train robots with. While the internet and low-cost crowd-source platform have enabled the collections of massive dataset that fuels the advances in computer vision and natural language tasks, such mechanism has yet to exist for robotic manipulation. Most datasets used in robotics paper are collected from constrained lab setting. The models trained from these datasets will often not generalize to realistic use cases. Third, robotics is just really hard, especially for manipulation tasks where the robot needs to act in complex ways to manipulate previously unseen objects. Despite commendable efforts by researchers, we still do not have in-depth physical understanding of the many aspects of the problems. For example, the study of friction, a central aspect of object manipulation, is still an active area of research \cite{is_tribology_approaching_its_golden_age_grand_challenges_in_engineering_education_and_tribological_research}. Even if learning algorithm does not use physical insights directly in their inference step, knowing the relevant physical model-based knowledge will allow researchers to develop better learning methods and debug them faster when they do not work.





\bibliographystyle{ACM-Reference-Format}
\bibliography{main}

\end{document}